%% file: main.tex
\pdfoutput=1
\documentclass[a4paper]{techreport}
\usepackage{tikz}
\usepackage{xurl}
\usetikzlibrary{arrows.meta,positioning,calc}
\graphicspath{{figures/}}
\newcommand{\method}{\textsc{Segment--Snap}\xspace}
\newcommand{\mao}{\ensuremath{\mathrm{AP}_{50}^{\mathrm{AO}}}\xspace}
\newcommand{\ap}{\ensuremath{\mathrm{AP}_{50}}\xspace}
\newcommand{\apaxis}{\ensuremath{\mathrm{AP}_{50}^{\mathrm{A}}}\xspace}
\newcommand{\aporigin}{\ensuremath{\mathrm{AP}_{50}^{\mathrm{O}}}\xspace}
\newcommand{\apmotion}{\ensuremath{\mathrm{AP}_{50}^{\mathrm{AO}}}\xspace}
\crefname{equation}{Eq.}{Eqs.}
\Crefname{equation}{Eq.}{Eqs.}
\crefname{section}{Sec.}{Secs.}
\Crefname{section}{Sec.}{Secs.}
\crefname{subsection}{Sec.}{Secs.}
\Crefname{subsection}{Sec.}{Secs.}
\crefname{table}{Tab.}{Tabs.}
\Crefname{table}{Tab.}{Tabs.}
\crefname{figure}{Fig.}{Figs.}
\Crefname{figure}{Fig.}{Figs.}
\colorlet{rptrow}{rptgreen5!15!white}
\colorlet{rpttablehead}{rptgreen5!45!white}
\newcommand{\tablehead}{\rowcolor{rpttablehead}}
\renewcommand{\arraystretch}{1.08}
\hypersetup{pdftitle={Geometric and Semantic Coupling for Interaction Understanding in 3D Scenes},pdfauthor={Hanyang Kong; Xingyi Yang}}

\title{Geometric and Semantic Coupling for\\Interaction Understanding in 3D Scenes}
\shorttitle{Coupled Part-Handle Reasoning}
\reporttype{Technical Report}
\author[1]{Hanyang Kong}
\author[2]{Xingyi Yang}
\affiliation[1]{National University of Singapore}
\affiliation[2]{The Hong Kong Polytechnic University}
\correspondence{Xingyi Yang (\email{xingyi.yang@polyu.edu.hk})}
\infoline[Author email]{Hanyang Kong (\email{hanyang.k@u.nus.edu})}
\date{September 27, 2026}
\projectpage{https://hyokong.github.io/segment-snap-page/}
\abstract{%
Understanding interaction in a 3D scene requires recovering movable parts, their motion, and where they can be operated. These quantities are related, and their predictions can inform one another.
A closed cabinet door, for instance, reveals a movable surface but may leave the hinge side ambiguous; its handle helps resolve this ambiguity, while the part provides context for localizing and interpreting the small handle.
Building on this observation, we present \method, which combines geometric and semantic evidence through part-handle coupling.
Three independently trained predictors recover movable parts, dense handles, and part-associated handle proposals. We couple their outputs in two directions.
For part motion, a training-free geometric decoder fits predicted part surfaces under explicit physical priors and uses detected handles to select candidate hinge lines.
For handle prediction, a part-conditioned branch proposes additional handles, while standalone part classes refine their rotation/translation labels, with dense-handle labels as a fallback. Each transfer is applied once, without iterative feedback.
On the Articulate3D validation set, handle guidance raises motion-gated AP from 13.74 to 40.98 under fixed masks and axes. Additional handle proposals raise handle AP from 24.63 to 29.65, and full contextual class correction raises it to 30.99 in the reference configuration.
Fixed-input controls, retraining ablations, learned-decoder comparisons, and paired visualizations together characterize the benefits and limits of this coupling. Our system also achieved first place in the Articulate3D Challenge.
}
\input{figures/teaser}

\begin{document}
\maketitle
\clearpage
\input{sections/01_introduction}
\input{sections/02_related}
\input{sections/03_method}
\input{sections/04_experiments}

\input{sections/05_discussion}
\FloatBarrier
\bibliographystyle{plainnat}
\bibliography{refs}
\clearpage
\appendix
\raggedbottom
\let\reportappendixsection\section
\renewcommand{\section}{\FloatBarrier\reportappendixsection}
\section*{Appendix: Additional Details and Results}
This appendix specifies the implementation and evaluation protocol (\Cref{app:implementation}),
additional quantitative controls and uncertainty (\Cref{app:controls}), coverage and failure
accounting (\Cref{app:coverage}), and qualitative analyses of both coupling directions
(\Cref{app:qualitative}). \Cref{app:external} details the REACT3D comparison, additional
scene examples, and complete-scene accuracy and cost summaries. All component experiments
use the public 42-scene validation split; the challenge standings are reported separately
in \Cref{sec:challenge}.
\input{sections/06_appendix}
\input{sections/07_external_comparison}

\end{document}

%% file: figures/teaser.tex
\teaser[fig:teaser]{\includegraphics[width=\linewidth]{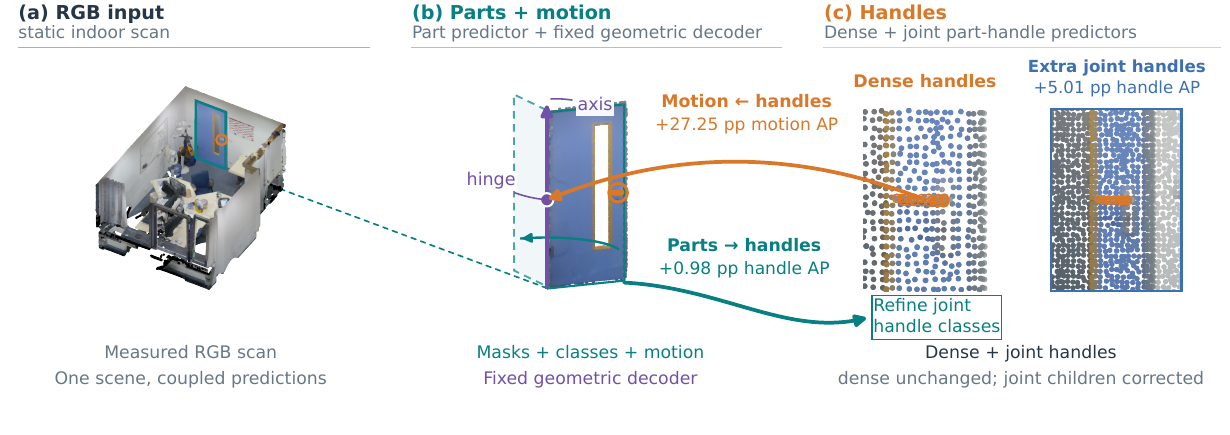}}{%
\textbf{Coupled interaction understanding from a static RGB point cloud.} Handles guide a training-free geometric decoder to select a part's hinge side (+27.25 percentage points in motion AP). Conversely, a joint part-handle predictor adds handles (+5.01 points in handle AP), while part classes refine their rotation/translation labels (+0.98 points). Orange and teal arrows show one-pass transfers without iterative feedback. Gains are from controlled validation comparisons (\Cref{tab:coupling}).}

%% file: sections/01_introduction.tex
\section{Introduction}
\label{sec:intro}

A 3D reconstruction can tell us that a cabinet is present without explaining how to open it. Progress in annotated indoor reconstruction and open-vocabulary scene representations makes objects increasingly accessible to semantic queries~\citep{dai2017scannet,yeshwanth2023scannetpp,takmaz2023openmask3d,gu2024conceptgraphs}. Interaction requires a richer description: which surfaces are movable, where they can be operated, and whether they rotate about a hinge or translate along a direction. Recovering this structure connects scene perception to the geometry needed for subsequent interaction. Recent benchmarks make these questions concrete by annotating movable parts, motion parameters, and small operating regions in real indoor scans~\citep{halacheva2025articulate3d,delitzas2024scenefun3d}. We call the operating regions \emph{handles}, including regions without a conventional handle shape. Our task is to infer this joint description from a static RGB point cloud, without observing an object move or executing an action.

Two coupled ambiguities make this task more demanding than object recognition. The first is \emph{scale and context}. A door occupies a broad surface, whereas its handle may comprise only a few observed points. Grouping points into coherent regions helps delineate the door but can absorb a small handle into its surroundings. Conversely, preserving fine spatial detail can reveal a handle while leaving its function ambiguous: a locally similar region may operate either a hinged door or a sliding drawer. Accurate localization therefore does not automatically give the correct motion class.

The second ambiguity is \emph{motion geometry}. A closed, approximately planar panel constrains the orientation and extent of a movable part, but not necessarily its attachment side. Even a perfect door mask can admit a vertical hinge on either edge. Observations across articulation states can expose that joint~\citep{jiang2022ditto,liu2023paris}, but this evidence is absent from a single static scene. A motion estimate must distinguish the alternatives using other cues. This matters because an instance may be well segmented yet fail a motion-sensitive metric, while a small but correctly located handle can provide the missing geometric evidence.

Existing approaches address these ambiguities through different sources of evidence. USDNet combines instance segmentation and dense motion prediction to recover movable parts, handles, and articulation, without explicitly using detected handle locations to select hinges~\citep{halacheva2025articulate3d}. Functional scene graphs relate interactive elements to objects but do not directly specify metric motion axes or origins~\citep{zhang2025openfungraph}. Reconstruction approaches infer joints from completed geometry or scene-to-simulation pipelines~\citep{zhao2025real2code,iliash2026s2o,huang2026react3d}. Our question is complementary: how can the part and handle predictions already available in a static scan provide evidence for one another?

We exploit the physical attachment between a handle and its movable part. On many upright doors, a handle lies away from the hinge, making its position informative about the attachment side. In the reverse direction, the surrounding part supplies two kinds of context: a larger region from which to propose a tiny handle, and a motion class with which to interpret that handle. The first concerns detection; the second concerns the label of an existing detection. Thus, \emph{handle locations guide part motion, while part-associated proposals and part classes improve handle prediction}. This relation connects the outputs without requiring the same spatial representation or inference rule for both.

We introduce \method, which separates scale-appropriate perception from explicit evidence transfer (\Cref{fig:teaser,fig:pipeline}). Three independently trained predictors have distinct responsibilities. A standalone \emph{part predictor} supplies the final movable surfaces and labels each as rotating or translating. A \emph{dense handle predictor} classifies individual points and groups them into initial handle instances. A \emph{joint part-handle predictor} supplies a second handle source: a learned feature representing each of its own parts predicts both that part and an associated handle. These internal parts organize the additional proposals; they do not replace the standalone parts. The design therefore preserves the standalone part and dense-handle predictions while allowing part-associated detections to supplement the handle set.

For motion, we fit the standalone part surfaces and construct candidate hinge lines under explicit planar and upright priors. An associated dense handle selects the candidate farthest from its location. Translating parts instead use the thin direction of their fitted geometry. Axes and origins are computed by a fixed geometric decoder rather than a trained motion-regression head. Only this decoder is training-free; the perception networks supplying its inputs are learned.

For handles, the joint predictor provides additional candidates, initially labelled by their internal parent parts. Confident standalone parts that contain these candidates can correct their rotation/translation labels; overlapping dense handles provide a fallback when no part qualifies. We append the resulting candidates without changing the original dense detections. This reverse transfer uses part context for both proposals and labels, but the two operations draw on different part predictors. Each transfer is applied once, without feeding the final handles back to recompute hinges.

On Articulate3D validation, handle guidance raises motion-gated AP from 13.74 to 40.98 while keeping part masks, classes, scores, and axes fixed. Appending associated handle proposals raises handle AP from 24.63 to 29.65. Contextual class correction of these added handles then raises AP to 30.99 in the reference configuration, with a smaller and more variable gain across repeated training. Our controls distinguish the benefit of an associated proposal source from that of query conditioning alone, and separate label correction from changes in localization.

The resulting system also ranked first for both movable-part motion and handle detection in the Articulate3D Challenge. We report these public standings separately from validation experiments in \Cref{sec:challenge}.

Our contributions are:
\begin{itemize}
\item A geometric and semantic coupling strategy for interaction understanding: dense handles guide part motion, while part-associated proposals and standalone part classes improve handle detection through distinct, one-pass transfers.
\item A training-free motion decoder that separates support fitting, physical axis priors, and handle-guided hinge selection, using independently predicted scene handles to resolve hinge-side ambiguity.
\item An evaluation combining fixed-input interventions, learned-decoder controls, repeated training, and scene-level comparisons, separating the effects of coupling from differences in inputs, supervision, and inference cost.
\end{itemize}

%% file: sections/02_related.tex
\section{Related work}
\label{sec:related}

\paragraph{From semantic to functional scene descriptions}
ScanNet and ScanNet++ support semantic and instance-level indoor scene understanding~\citep{dai2017scannet,yeshwanth2023scannetpp}; 3DSSG and ConceptGraphs add semantic and open-vocabulary object relations~\citep{wald2020semanticscenegraphs,gu2024conceptgraphs}. SceneFun3D annotates functionality and affordances~\citep{delitzas2024scenefun3d}, while Articulate3D combines movable parts, interaction regions, and motion parameters, jointly predicted by USDNet~\citep{halacheva2025articulate3d}. OpenFunGraph relates interactive elements to objects~\citep{zhang2025openfungraph}. We use the part-handle relation to inform predictions: associated handles select metric hinge hypotheses, while part evidence supplements handle proposals and labels.

\paragraph{Point-cloud representations and instance prediction}
PointNet++ and Point Transformer V3 learn point-cloud features through hierarchical neighborhoods and serialized attention, respectively~\citep{qi2017pointnetplusplus,wu2024pointtransformerv3}; superpoint methods organize context around coherent regions~\citep{landrieu2018superpointgraphs,robert2023superpointtransformer}. PointGroup and SoftGroup group pointwise evidence into instances~\citep{jiang2020pointgroup,vu2022softgroup}, whereas Mask3D, SPFormer, and OneFormer3D decode masks with queries~\citep{schult2023mask3d,sun2023spformer,kolodiazhnyi2024oneformer3d}. OpenMask3D adds open-vocabulary features, and Open3DIS uses multi-view 2D mask guidance~\citep{takmaz2023openmask3d,nguyen2024open3dis}. We adopt established perception components with broad region support for parts and fine point/voxel support for handles; our contribution is their coupling.

\paragraph{Parts, affordances, and interaction}
PartNet supplies hierarchical part annotations, SAPIEN provides articulated simulation assets, and PartSLIP transfers image-language knowledge to low-shot part segmentation~\citep{mo2019partnet,xiang2020sapien,liu2023partslip}. Where2Act predicts push/pull actionability; GAPartNet connects actionable part categories to cross-category perception and manipulation~\citep{mo2021where2act,geng2023gapartnet}. We predict an interaction description, not an action policy. A handle's rotation/translation label describes the part it operates, motivating contextual correction when local handle appearance is ambiguous.

\paragraph{Articulation from static and changing observations}
Shape2Motion estimates movable parts and motion from static 3D shapes; ANCSH recovers articulated pose and joints from a depth point cloud~\citep{wang2019shape2motion,li2020categorylevel}. OPD and OPDMulti predict openable parts and motion from single images~\citep{jiang2022opd,sun2024opdmulti}. Ditto instead uses observations before and after interaction, while PARIS uses two articulation states~\citep{jiang2022ditto,liu2023paris}. We infer motion from a static scene using learned part-handle evidence and explicit planar/upright priors, without a separate motion regressor or observed state changes.

\paragraph{Reconstruction and geometric articulation priors}
Real2Code reconstructs articulated objects and generates joint code; Articulate-Anything uses a vision-language model to construct articulated assets~\citep{zhao2025real2code,le2025articulateanything}. S2O adds openable parts and interiors to static container meshes~\citep{iliash2026s2o}; its \href{https://github.com/3dlg-hcvc/s2o/blob/35aa03f29b856f123c476c006638a38fc1e3ad4c/opmotion/engine/motion_predictor/rule_motion_predictor.py}{released rule} already selects box edges opposite a geometry-derived handle proxy. We do not claim this primitive as new. REACT3D targets scene-to-simulation recovery with articulation and completion~\citep{huang2026react3d}. Independently detected scene handles guide our motion decoding, while part-associated proposals and standalone part classes improve the handle output. Comparisons retain the differences in inputs and training conditions.

%% file: sections/03_method.tex
\section{Method}
\label{sec:method}

\method separates learning to locate parts and handles from reasoning about their physical relationship. Three independently trained predictors provide final movable-part instances, dense handle detections, and additional part-associated handle proposals. Their outputs meet in two directed transfers: dense-handle locations guide a training-free decoder of part motion, while joint-model proposals augment handle coverage and standalone part classes refine their labels. The decoder has no learned motion head; only the perception networks are trained. We first define these prediction interfaces, then describe their learning and the two transfers. \Cref{fig:pipeline} shows the complete workflow.

\subsection{Problem formulation and prediction interface}
\label{sec:predictors}

The input is a static cloud of $N$ points, $X=\{(\boldsymbol x_n,\boldsymbol c_n,\boldsymbol n_n)\}_{n=1}^N$, where $\boldsymbol x_n$, $\boldsymbol c_n$, and $\boldsymbol n_n$ are the 3D position, measured RGB, and surface normal of point $n$. Coordinates are expressed in an upright scene frame. Each output movable part has an instance mask, confidence, rotation/translation class, unit motion axis, and representative origin. Each handle has a mask, confidence, and the motion class of the part it operates. The objective is to predict these quantities from the observed scan, without motion sequences, action execution, or completed object geometry.

Three networks process the same input using the same backbone architecture but separate parameters and independent training. Their outputs are coupled only at inference:
\begin{equation}
f_{\mathrm{part}}(X)=\mathcal P,\qquad
f_{\mathrm{dense}}(X)=\mathcal H_d,\qquad
f_{\mathrm{joint}}(X)=(\mathcal Q,\mathcal H_c).
\label{eq:predictors}
\end{equation}
The standalone parts $\mathcal P=\{(M_i,s_i,y_i)\}$ supply \emph{all final movable-part masks and classes}: $M_i$ is a set of input-point indices, $s_i\in[0,1]$ its confidence, and $y_i\in\{\mathrm{rotation},\mathrm{translation}\}$ its motion class. They also provide the surfaces for geometric fitting and the context for handle-label correction. Dense handles $\mathcal H_d$ are both initial handle detections and spatial cues for motion.

The joint predictor supplies additional handles $\mathcal H_c$ associated with its \emph{own} parts $\mathcal Q$. We call each such part and its handle a parent-child pair. Both are predicted from a common part-query feature within this network; the child is not predicted from a thresholded parent mask. Each parent or handle prediction carries a mask, confidence, and class. Although $\mathcal P$ and $\mathcal Q$ describe the same kind of movable part, their roles differ: $\mathcal P$ determines final part outputs and contextual labels, whereas $\mathcal Q$ associates and initializes the additional handle proposals. The joint model neither takes $\mathcal P$ as input nor replaces it. No matching between the two part sets is required; standalone parts influence children through spatial containment.

\input{figures/pipeline}

\subsection{Learning broad parts and fine interaction regions}
\label{sec:perception}

A movable surface and its operating region require different spatial support. Superpoint grouping offers a compact representation of a broad door, but a handle occupying only a few points can disappear under majority pooling. We therefore use instance queries for parts and fine point/voxel masks for handles. Each network has its own Volt-B voxel Transformer initialized from ScanNet++ pretraining~\citep{yilmaz2026volt}, with 2\,cm voxels and RGB/normal features. This common backbone supports different prediction heads; the contribution lies in combining their outputs, not in a new backbone.

\paragraph{Standalone part instances}
An SPFormer decoder~\citep{sun2023spformer} attends to features pooled over graph-partitioned superpoints~\citep{felzenszwalb2004efficient}. Its 200 queries are learned instance features, each predicting a part mask, rotation/translation class, and confidence score. Hungarian assignment supervises the queries using classification, mask binary cross-entropy (BCE), Dice, and score terms. At inference, each standalone query emits its highest-scoring class before instance filtering. This yields $\mathcal P$ without treating different class hypotheses from one query as different parts. The network does not regress motion axes or origins.

\paragraph{Dense handle instances}
A separate network predicts background, rotation-handle, and translation-handle probabilities at each point. Class-weighted cross-entropy with label smoothing and multiclass Lov\'asz loss supervise the field~\citep{berman2018lovasz}. Maximum-probability labels are grouped by class using a 2.5\,cm-radius graph. Components with at least three points become instances, scored by their mean assigned-class probability. The small minimum size and fine support retain candidates that a part-scale grouping could suppress. These detections remain unchanged in the final handle set.

\paragraph{Joint parent-child prediction}
The joint network first decodes its own part-query features from its backbone. Each query $\boldsymbol q_j$ then feeds a parent-part head and a child-handle head in parallel. The heads share this instance feature within the joint network; the child head does not wait for, crop by, or consume a thresholded parent mask. Instead, it compares the part-query feature with fine-resolution voxel features. For voxel $v$ with feature $\boldsymbol f_v$, the child-head output $p_{jv}$ is the probability of belonging to query $j$'s handle:
\begin{equation}
p_{jv}=\operatorname{sigmoid}\!\left(g_q(\boldsymbol q_j)^\top g_v(\boldsymbol f_v)\right),
\qquad
\mathcal L_{\mathrm{child}}
=\mathcal L_{\mathrm{BCE}}+\mathcal L_{\mathrm{Dice}}+2\mathcal L_{\mathrm{Tv}}.
\label{eq:child}
\end{equation}
Here $g_q,g_v$ project query and voxel features to a common dimension. The child loss combines BCE, Dice, and Tversky ($\mathcal L_{\mathrm{Tv}}$) terms. Parent-only Hungarian assignment determines the ground-truth part for each query; the child target is its associated interactable region. A voxel is positive if it contains any target child point. Tversky false-positive and false-negative weights are 0.7 and 0.3~\citep{salehi2017tversky}, penalizing the spread of a small handle over its parent surface. Parent and child heads are trained together within this third network, independently of the standalone networks. Training and selection details are in \Cref{app:implementation}.

\subsection{Handles to parts: geometric motion decoding}
\label{sec:geometry}

A part surface constrains plausible motion but may leave the attachment side ambiguous. We separate this inference into reliable support fitting, a physical axis prior, and handle-guided hinge selection. In particular, a door's extent supplies candidate hinge locations, while an associated handle distinguishes the two sides. The decoder assumes approximately planar surfaces and upright rotational axes; it applies fixed geometric rules, not a learned model of arbitrary articulation.

\paragraph{Reliable support and a local frame}
Detached points can distort a box fit even when the predicted mask remains useful. For each $M_i$, we fit geometry only to its largest connected component $S_i$ under a 5\,cm-radius graph. The output mask and class remain unchanged. We also recalibrate confidence to $\bar s_i=s_i|S_i|/|M_i|$, where $|\cdot|$ counts points, downweighting fragmented predictions. Fitting affects geometry; rescoring affects ranking.

We estimate a plane from $S_i$ by principal component analysis (PCA) and fit a minimum-area rectangle over the projected convex hull. This supplies orthonormal directions $\boldsymbol b_1,\boldsymbol b_2,\boldsymbol b_3$ and extents $\ell_1\geq\ell_2\geq\ell_3$. We anchor the candidate construction at the mean $\boldsymbol\mu_i$ of the points used for fitting, not at the rectangle's bounding-box center. Fitting uses at most 20,000 deterministically sampled support points. Suppressing part indices on the local directions and extents, we assign the unit motion axis
\begin{equation}
\hat{\boldsymbol a}_i=
\begin{cases}
\boldsymbol e_z,&y_i=\mathrm{rotation},\\
\boldsymbol b_3,&y_i=\mathrm{translation}.
\end{cases}
\label{eq:axis}
\end{equation}
Here $\boldsymbol e_z=(0,0,1)^\top$ is the scene's vertical unit vector. For an upright door, the hinge direction is fixed in that frame. For a drawer, the front's thin direction approximates translation normal to the surface; its representative origin is $\boldsymbol\mu_i$. The remaining handle-guided steps apply only to rotation-class parts. A handle cue selects a rotational origin; it cannot repair an incorrect axis prior.

\paragraph{Associating a handle and constructing candidates}
For each dense-handle instance, we compute its point centroid. For part $i$, we select the single centroid $\boldsymbol h$ with the smallest distance to its support, where $d(\boldsymbol h,S_i)=\min_{n\in S_i}\|\boldsymbol h-\boldsymbol x_n\|_2$, and accept it only if $d(\boldsymbol h,S_i)<0.5$\,m. Association uses handle location, not confidence or rotation/translation label: geometric usefulness and handle classification accuracy are distinct properties.

Let $r,s$ index the two box directions least aligned with $\hat{\boldsymbol a}_i$ in absolute dot product. Four anchors define candidate lines parallel to that axis:
\begin{equation}
\boldsymbol v_{\alpha\beta}=\boldsymbol\mu_i+
\tfrac{\alpha\ell_r}{2}\boldsymbol b_r+
\tfrac{\beta\ell_s}{2}\boldsymbol b_s,
\qquad \alpha,\beta\in\{-1,1\}.
\label{eq:candidates}
\end{equation}
Index these anchors as $\boldsymbol v_k$, $k=1,\ldots,4$; candidate $k$ is the line through $\boldsymbol v_k$ in direction $\hat{\boldsymbol a}_i$. For a thin door they represent two attachment sides, with a front/back pair on each. These mean-anchored, box-derived lines need not coincide with literal rectangle edges.

\paragraph{Selecting the hinge}
On many doors the handle lies away from the hinge. We therefore use $\boldsymbol h$ to select the candidate line farthest from the handle, then project the support mean onto that line to obtain a representative origin. Let $\Pi_i=I-\hat{\boldsymbol a}_i\hat{\boldsymbol a}_i^\top$ project onto the plane perpendicular to the axis, where $I$ is the $3\times3$ identity. The selection and projection are
\begin{equation}
k^*=\arg\max_k\|\Pi_i(\boldsymbol h-\boldsymbol v_k)\|_2,\qquad
\hat{\boldsymbol o}_i=\boldsymbol v_{k^*}+
\hat{\boldsymbol a}_i\hat{\boldsymbol a}_i^\top(\boldsymbol\mu_i-\boldsymbol v_{k^*}).
\label{eq:origin}
\end{equation}
The origin is a representative point on a hinge line, not a uniquely identified physical pivot. Without an accepted handle it falls back to $\boldsymbol\mu_i$. Translation evaluation has no origin clause. Opposite-handle reasoning has precedent in S2O~\citep{iliash2026s2o}; here its evidence comes from an independently detected scene handle. All fitting and selection operations are fixed. Consequently, the \emph{motion decoder} is training-free even though the perception networks supplying its inputs are learned.

\subsection{Parts to handles: complementary proposals and labels}
\label{sec:handles}

The reverse transfer has two distinct steps. The joint predictor uses its own part features to propose handles (\Cref{sec:perception}); standalone part predictions then provide a separate source of evidence for their motion labels. We preserve the association that produced each proposal and correct only its class, leaving the original dense detections unchanged.

\paragraph{Preserving parent-child association}
After filtering and reordering the joint model's parent detections, we retain each surviving parent's original query index and use it to retrieve the corresponding child field. Association therefore follows the shared query, not output position or a new spatial match. The child mask contains points whose voxel probability exceeds 0.30. It is neither split into components nor clipped to its parent, so an imperfect parent boundary need not truncate an otherwise localized handle. Its initial class and score come from that parent, with confidence scaled by 0.05. Children are appended as additional detections, not merged with dense masks; the original dense masks, classes, and scores are retained.

This union tests whether an additional, correctly associated source recovers missing handles. It does not by itself isolate query conditioning as the cause of the gain; the matched conditioning and association controls are separated in \Cref{app:controls}. No monotonic-AP guarantee is assumed for arbitrary added proposals.

\paragraph{Semantic correction from containing parts}
A child can be well localized but inherit the wrong motion class from its joint-model parent. The standalone parts supply a separate contextual prediction. For child mask $H$, an eligible part satisfies
\begin{equation}
\frac{|H\cap M_i|}{|H|}\geq0.9,\qquad \bar s_i\geq0.3.
\label{eq:containment}
\end{equation}
Here $\bar s_i$ is the component-rescaled confidence from \Cref{sec:geometry}. We first select the highest-confidence eligible part and assign its class to the child. Only if no part qualifies do we consult dense handles, using the one with the largest child containment and requiring at least 90\% containment. If neither source qualifies, the child keeps its initial parent label. A qualifying part takes priority even when its class already agrees with the child; the fallback is not an equal vote between sources. Only child classes change; their geometry and confidence, and every dense detection, remain fixed.

\paragraph{Final outputs and inference schedule}
The final part output is $\{(M_i,\bar s_i,y_i,\hat{\boldsymbol a}_i,\hat{\boldsymbol o}_i)\}$, and the final handle output concatenates $\mathcal H_d$ with the class-corrected children. Each information transfer is applied once. Dense handles guide hinges; standalone parts guide child labels. Decoded axes and origins are not inputs to class correction, and the final handle union is not fed back into motion decoding. This directed computation makes each source of evidence independently testable.

%% file: figures/pipeline.tex
\begin{figure}[!t]
\centering
\includegraphics[width=\linewidth]{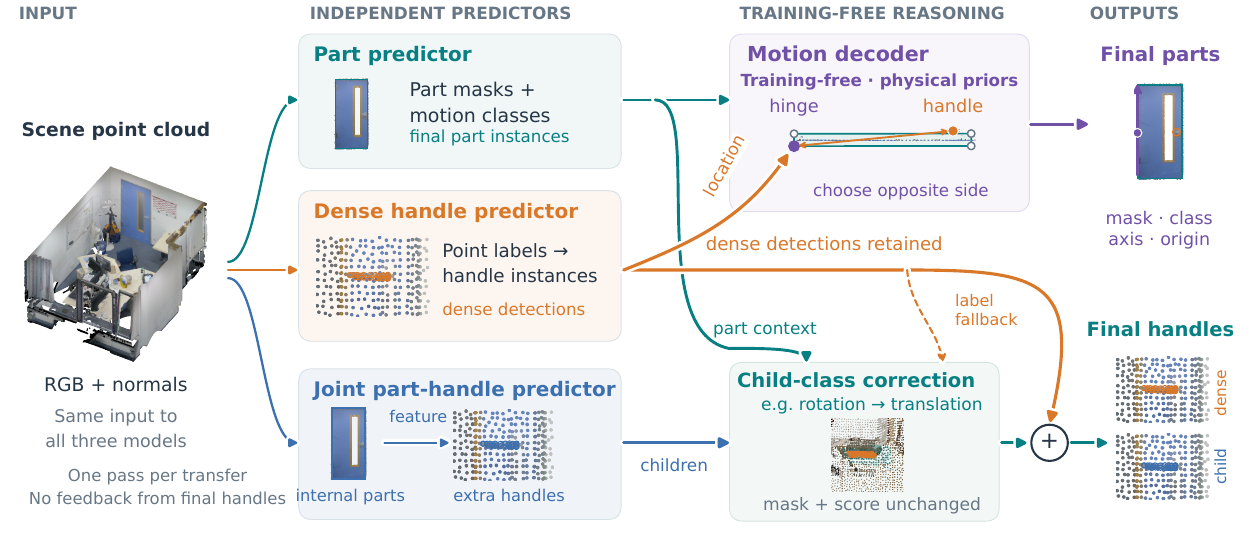}
\caption{\textbf{Three predictors, two directed transfers.} The same RGB point cloud and normals enter independently trained part, dense-handle, and joint part-handle predictors. Standalone parts supply all final part masks and classes. Dense handles provide locations to a training-free motion decoder, which fits part surfaces, applies explicit physical priors to the axes, and selects hinges opposite the handles. The joint predictor uses its own part features to propose additional handles. Standalone parts correct child classes, with dense labels as fallback, before the children are appended to unchanged dense detections. Only the added handles' classes are corrected; their masks and scores remain fixed. The final handles do not feed back into motion decoding.}
\label{fig:pipeline}
\end{figure}

%% file: sections/04_experiments.tex
\section{Experiments}
\label{sec:experiments}

We evaluate interaction understanding as a joint output of movable parts, motion, and handles. We first place the reference system in published benchmark context, then compare articulated-part outputs on the same validation scenes, and finally isolate the two directed transfers with fixed-input controls.

\subsection{Protocol and published comparison}
\label{sec:evaluation-protocol}

All perception models are trained on Articulate3D's 195 training scenes and evaluated on its public 42-scene validation split~\citep{halacheva2025articulate3d}. Model and hyperparameter selection consulted validation data, so these are development-set comparisons rather than an untouched generalization estimate. We report macro AP at IoU 0.5 for parts and handles. Axis-gated part AP additionally requires a sign-invariant axis error below $15^\circ$. Joint motion AP (\apmotion) also tests rotational origins: each representative origin must lie within 0.25\,m of the other axis line. Translations have no origin clause. The stricter Euclidean-origin variant is reported separately only where needed for comparison with published results. Scores are percentages and changes are percentage points (pp).

\input{tables/comparison}

\Cref{tab:comparison} compares the reference configuration with SoftGroup~\citep{vu2022softgroup}, Mask3D~\citep{schult2023mask3d}, and USDNet using values reported in the USDNet benchmark paper. \method reaches 47.93 part AP and 40.98 motion AP, compared with USDNet's 41.8 and 25.0; handle AP is comparable at 30.99 versus 31.1. These published numbers provide context, rather than a matched-resource experiment. The controlled and same-scene comparisons below establish the evidence for our system.

\subsection{Same-scene, common-output comparison}
\label{sec:external}

\input{tables/react3d_accuracy}

We run REACT3D~\citep{huang2026react3d} on all 42 validation scenes and evaluate a common output: part instances on the evaluator's point cloud, each with a motion class, axis, and origin. REACT3D's predicted meshes are mapped to the point cloud with a 2\,cm nearest-neighbour tolerance, and axes are normalized before evaluation. Three scenes with verified zero detections remain in the evaluation as empty prediction sets. REACT3D receives RGB-D keyframes, camera poses, and a mesh, whereas \method receives an RGB point cloud. Our predictors use the Articulate3D training split; REACT3D is not trained on it. Thus, this comparison evaluates outputs under different input and training conditions, not the superiority of one input modality.

On the common-output protocol, \method reaches 40.98 motion AP versus 8.09 for REACT3D (\Cref{tab:react3d-accuracy}); the paired scene-jackknife 95\% interval for the 32.89-pp difference is $[18.49,47.29]$. It describes scene-sampling uncertainty, not training variability. Coverage explains much of REACT3D's translation deficit: 98 of 154 translating parts have no overlapping prediction, whereas all seven mask-matched translations pass the axis gate. Among 61 mask-matched rotations, 18 fail only the origin test.

\input{figures/react3d_main}

\paragraph{Qualitative comparison}
\Cref{fig:react3d-main} compares part extents and motion in shared views. In the first scene, REACT3D assigns incorrect motion types and produces unmatched detections; the second also exposes extent and motion-parameter errors. The displayed \method predictions recover more valid part-motion instances but still miss annotated parts. These favorable examples are selected by a disclosed displayed true-positive advantage; they illustrate failure modes rather than estimate their frequency.

\input{tables/react3d_cost}

\paragraph{Inference cost}
\Cref{tab:react3d-cost} measures one complete inference pass per scene for articulated-part output. Our path includes superpoint extraction, the standalone part and dense-handle networks, and geometric motion decoding; the third predictor for additional handles is not required for this output. REACT3D starts from extracted keyframes, poses, and mesh and traverses 13 stages. Its median time is 3,638.1\,s per scene versus 33.78\,s for \method, and its as-shipped peak host memory is 22.94 versus 4.70\,GiB. Unequal GPU logging coverage precludes a matched GPU-memory ranking. \Cref{app:external} details the timing scope, memory refit, and logging coverage.

\subsection{What coupling contributes}

\input{tables/coupling}

Holding predicted part masks, labels, scores, and axes fixed, replacing centroid origins with handle-guided hinge lines raises motion AP from 13.74 to 40.98 (+27.25 pp). Part AP and axis-gated AP remain 47.93 and 43.75, respectively, so the gain isolates hinge placement rather than segmentation or axis prediction. It is concentrated on rotational parts: their motion AP rises from 1.88 to 56.37, while translations are unchanged because their origins are not evaluated.

The reverse transfer adds complementary handle evidence. Appending associated children raises dense-handle AP from 24.63 to 29.65 (+5.01 pp). With masks and scores fixed, parts-only labels yield 30.63 (+0.98 pp), and the full rule including dense fallback reaches 30.99 (+1.34 pp). These label gains overlap and are not additive. Spatially permuting child locations gives no gain over 24.63, while corrupting part labels reduces corrected AP to 28.62 or 28.90. Thus, both localized proposals and meaningful class evidence matter. The proposal gain persists across child-model repeats; contextual correction is smaller and more variable (\Cref{tab:coupling-robustness}).

\subsection{Further controlled analysis}

\input{tables/validation_ablations}

\paragraph{Training-free motion decoding}
The decoder combines a world-$Z$ rotational-axis prior, a thin-box translation direction, and handle-guided hinge selection. On frozen part inputs, discrete learned selection has higher point estimates than continuous regression when both receive the same handle cue (\Cref{tab:motion-alternatives}). The training-free rule reaches 40.98, compared with at most 39.36 for the displayed learned heads on the reference part model. The two-part-model comparison in \Cref{tab:app-learned-two-bases} retains this ordering, but its paired intervals include zero. These results support an effective decoder without motion-regression training, not a statistically resolved or general advantage over learned motion estimation.

\input{tables/mechanisms}

\Cref{tab:geometry-controls} measures components both when added to a naive pipeline and when removed from the full decoder. The complete configuration improves motion AP from 12.24 to 40.98, with a paired gain interval of $[+22.12,+35.37]$ pp. Support cleanup is most useful with handle-guided geometry: it changes the fitted box and candidate lines, costs 2.95 pp when removed from the full decoder, but gives a $-0.11$-pp change when added alone. Isolated component gains therefore cannot simply be added.

\input{tables/semantic}

\paragraph{Complementary proposals and contextual labels}
The proposal controls in \Cref{tab:handle-complementarity} test whether the extra source contributes localized handles rather than merely extending the ranked detection list. Spatially permuted and random size-matched children return the dense-only AP. Real children preserve the dense detections' matching prefix and recover 53 additional GT handles, increasing AP by 5.01 pp. A separate conditioning/readout control does not isolate a reliable benefit of parent conditioning itself, so we attribute the gain to the implemented proposal source rather than conditioning alone (\Cref{tab:app-association}).

\input{tables/contextual_labels}

\Cref{tab:context-controls} isolates the source of child-label evidence. Part context and dense-incumbent context each improve AP by about one point; their combination reaches 30.99. Corrupting only the part-class channel instead reduces AP below the uncorrected union, despite changing more labels. Useful semantic evidence, rather than the number of corrections, explains the benefit. All rows preserve child masks and scores and leave dense detections unchanged.

\input{tables/robustness}

The three reference gains have positive paired scene intervals (\Cref{tab:coupling-robustness}). Handle guidance remains beneficial across four part-model draws. Appending children improves handle AP by 3.94--5.01 pp across three child-model draws, and its reference scene interval is $[+1.47,+8.56]$ pp; the gain is positive under every single-scene omission. Contextual correction also improves all three draws, but varies more relative to its gain, from 0.19 to 1.34 pp. These source-configuration comparisons test the listed predictors, not an isolated pretraining effect or a universal relationship between detector AP and hinge quality.

\subsection{Qualitative analyses}

\input{figures/qualitative_hinge}
\input{figures/qualitative_handles}

\Cref{fig:qual-hinge} holds each predicted mask fixed while comparing decoded origins. The paired successes show how handle location resolves the hinge side; non-vertical axes, missing handles, and residual errors expose the decoder's physical limits. \Cref{fig:qual-handles} shows complementary child detections and contextual label changes. Its over-extended mask and unmatched relabelled child illustrate why correct class context cannot repair poor localization. Additional galleries in \Cref{sec:appendix-visual} connect component ablations to changed geometric support, compare weaker prediction sources, and illustrate the failure census.

\FloatBarrier
\subsection{Public challenge result}
\label{sec:challenge}

Our Articulate3D challenge submission (team \emph{TnG}) achieved the highest published scores for movable-part motion and handle detection: 48.28\% \mao\ and 34.46\% \ap, respectively. \Cref{tab:leaderboard-mov,tab:leaderboard-handle} reproduce every public team entry and every metric exposed by the official board as accessed on September 6, 2026. These are challenge standings, not matched-resource comparisons. Test annotations are not public; all component studies above use public validation.

\input{tables/leaderboard}

%% file: tables/comparison.tex
\begin{table}[H]
\centering
\begin{threeparttable}
\caption{\textbf{Articulate3D validation results (\%).} Baseline values are reported in the USDNet benchmark paper~\citep{halacheva2025articulate3d}. Origin and axis columns add the corresponding motion gates to part AP$_{50}$.}
\label{tab:comparison}
\small
\setlength{\tabcolsep}{4pt}
\begin{tabularx}{\linewidth}{@{}Xrrrrr@{}}
\toprule
\tablehead
\textbf{Method} & \shortstack{\textbf{Part}\\\textbf{AP$_{50}$}} & \textbf{+Origin} & \textbf{+Axis} & \shortstack{\textbf{+Origin}\\\textbf{+Axis}} & \shortstack{\textbf{Handle}\\\textbf{AP$_{50}$}} \\
\midrule
SoftGroup & 32.7 & 18.5 & 21.5 & 17.7 & 14.5 \\
Mask3D & 39.1 & 24.4 & 33.8 & 19.3 & 30.2 \\
USDNet & 41.8 & 31.4 & 34.6 & 25.0 & \textbf{31.1} \\
\rowours \method & \textbf{47.93} & \textbf{43.11} & \textbf{43.75} & \textbf{40.98} & 30.99 \\
\bottomrule
\end{tabularx}
\end{threeparttable}
\end{table}

%% file: tables/react3d_accuracy.tex
\begin{table}[H]
\centering
\begin{threeparttable}
\caption{\textbf{Articulated-part results on all 42 validation scenes (AP, \%).} REACT3D receives RGB-D keyframes, camera poses, and a mesh, whereas \method uses an RGB point cloud. Both systems are evaluated in the same point-cloud domain. The joint +Origin +Axis metric follows \Cref{tab:comparison}; class-specific results are also shown.}
\label{tab:react3d-accuracy}
\small
\setlength{\tabcolsep}{3.5pt}
\begin{tabularx}{\linewidth}{@{}Xlrrrr@{}}
\toprule
\tablehead
& & & \multicolumn{3}{c}{\textbf{+Origin +Axis}} \\
\tablehead
\textbf{Method} & \textbf{Input} & \textbf{Part AP$_{50}$} & \textbf{All} & \textbf{Rotation} & \textbf{Translation} \\
\midrule
REACT3D & RGB-D, poses, mesh & 14.46 & 8.09 & 12.29 & 3.89 \\
\rowours \method & RGB point cloud & \textbf{47.93} & \textbf{40.98} & \textbf{56.37} & \textbf{25.59} \\
\bottomrule
\end{tabularx}
\end{threeparttable}
\end{table}

%% file: figures/react3d_main.tex
\begin{figure}[H]
\centering
\setlength{\tabcolsep}{1.5pt}
\renewcommand{\arraystretch}{1}
\begin{tabular}{@{}cccc@{}}
\small RGB & \small GT + axes & \small \method & \small REACT3D \\[2pt]
\includegraphics[width=.237\linewidth]{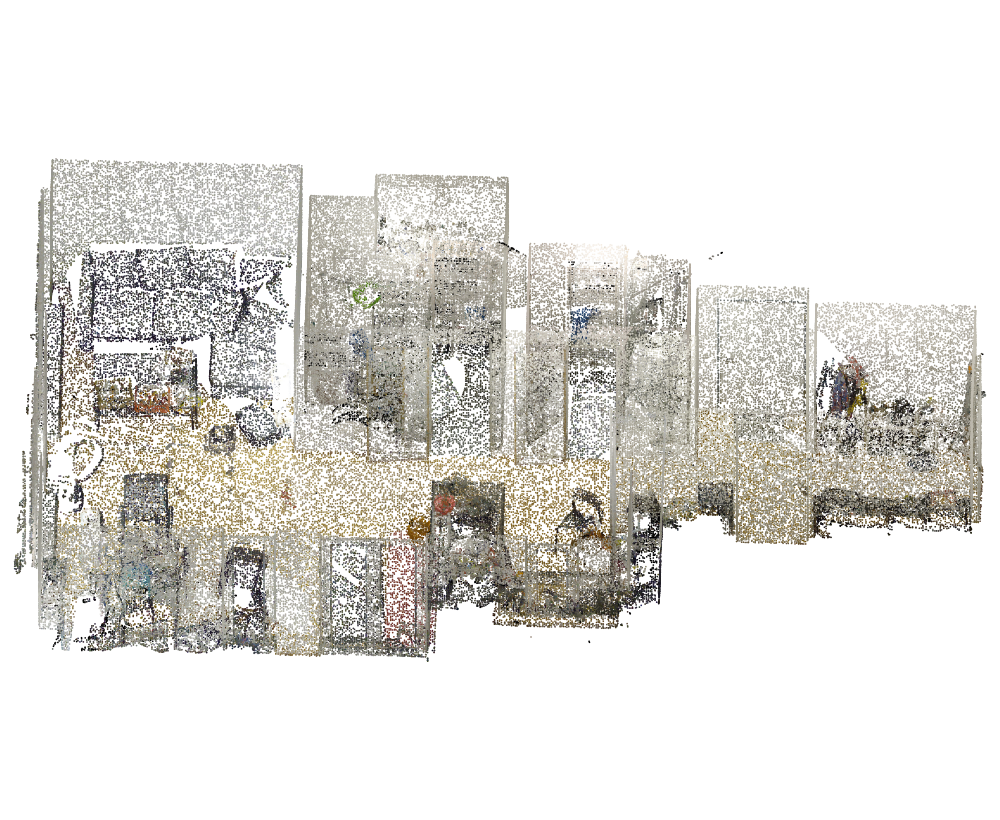} &
\includegraphics[width=.237\linewidth]{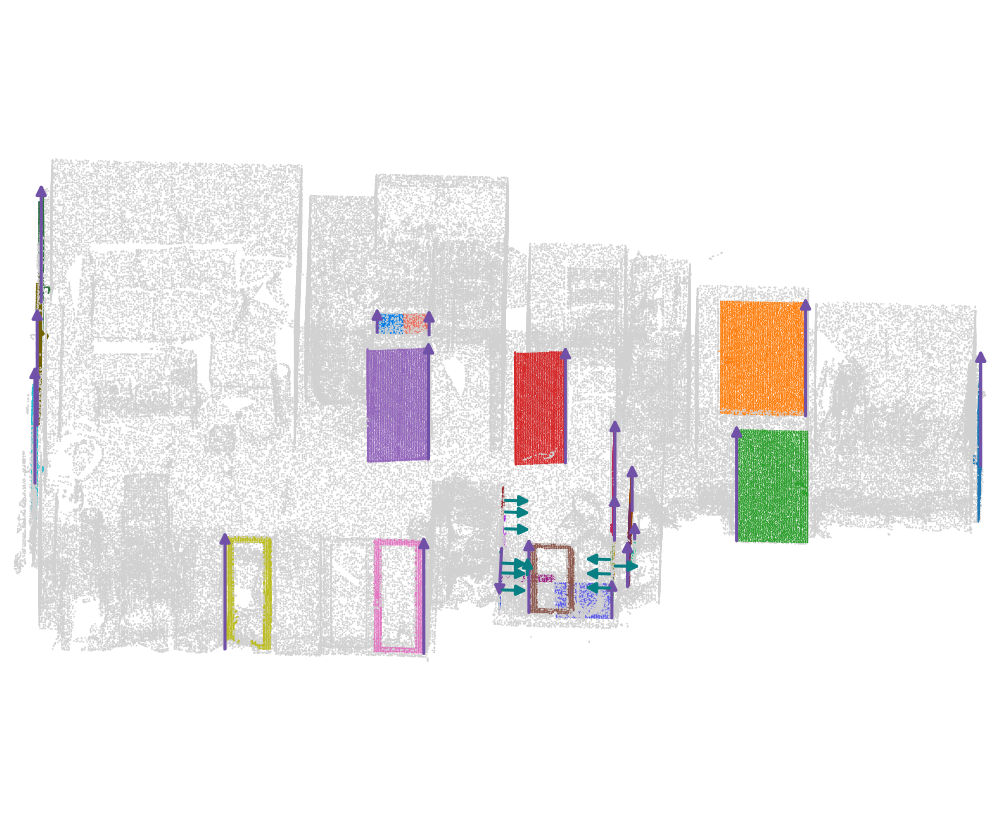} &
\includegraphics[width=.237\linewidth]{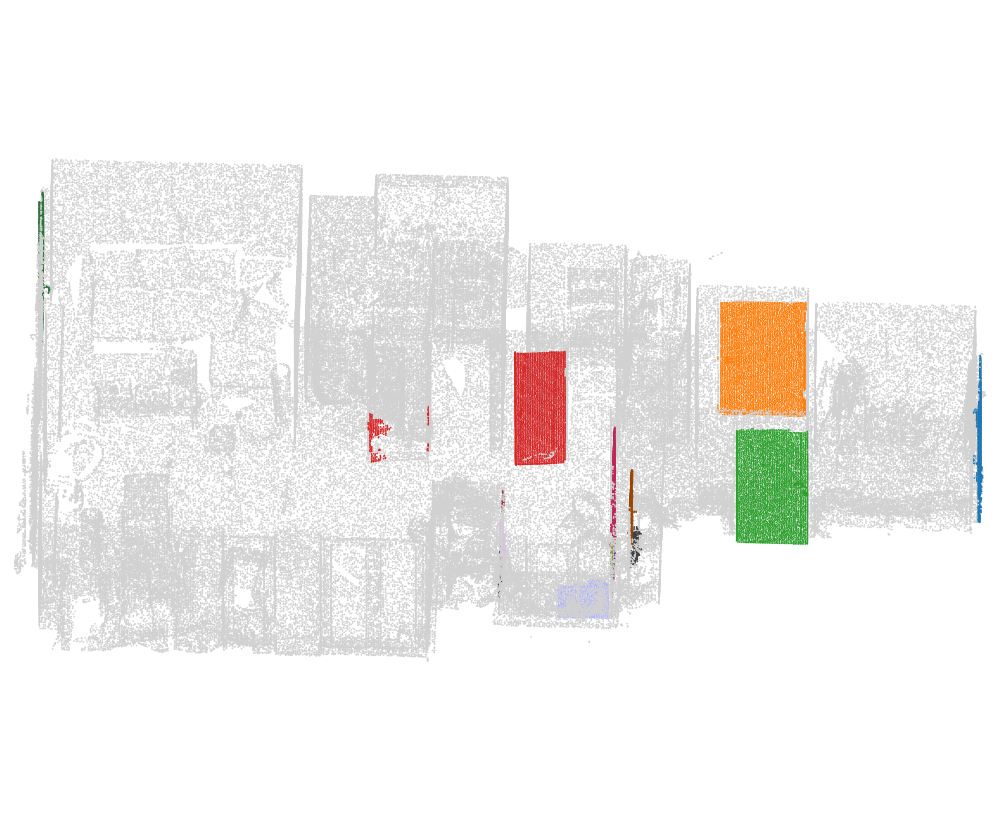} &
\includegraphics[width=.237\linewidth]{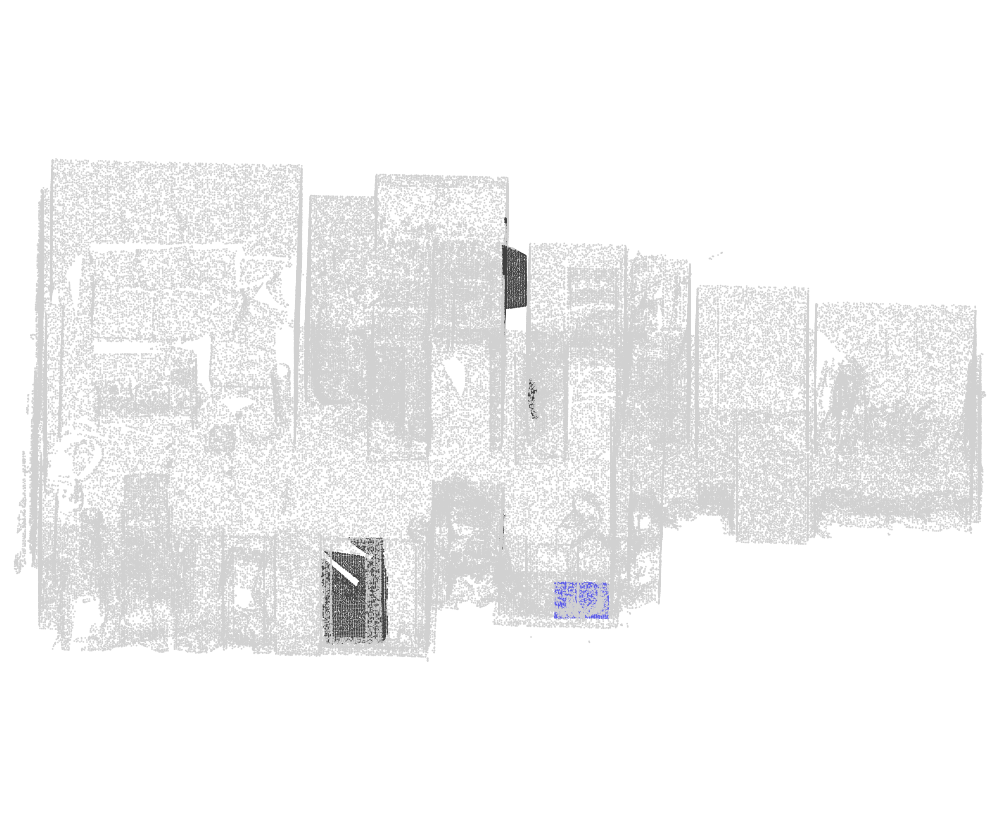} \\
\multicolumn{4}{c}{\scriptsize 61adeff7d5: 31 GT parts; ours 10 matched of 15 shown; REACT3D 1 matched of 5} \\[1pt]
\includegraphics[width=.237\linewidth]{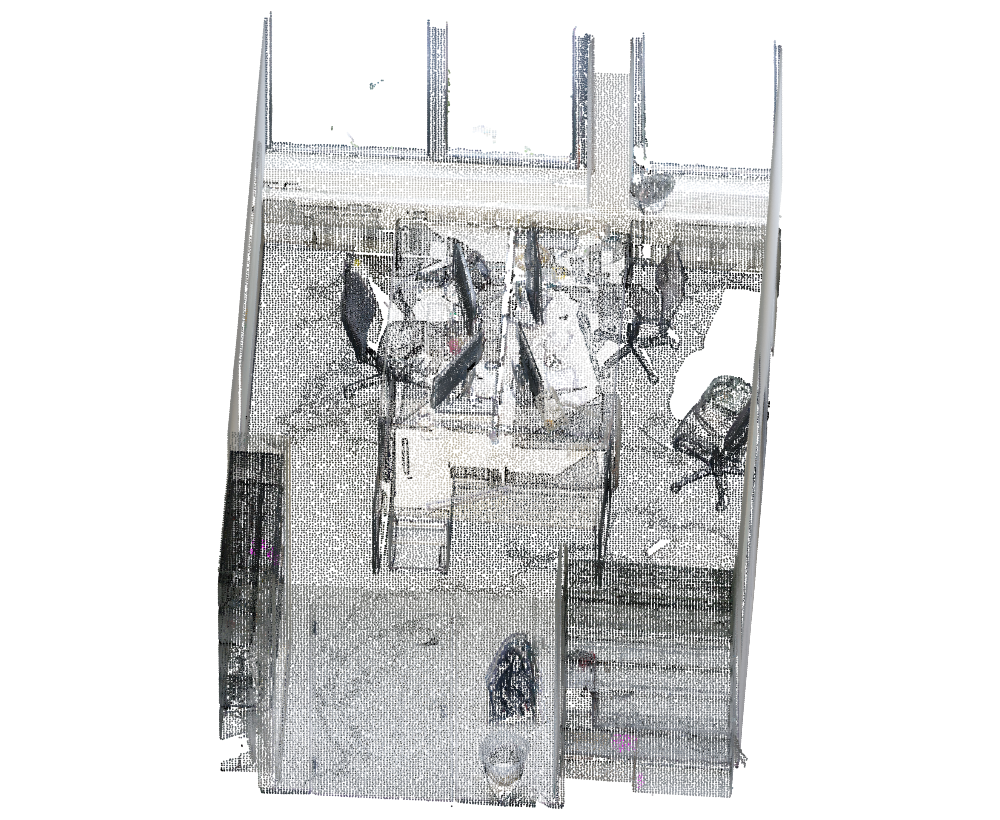} &
\includegraphics[width=.237\linewidth]{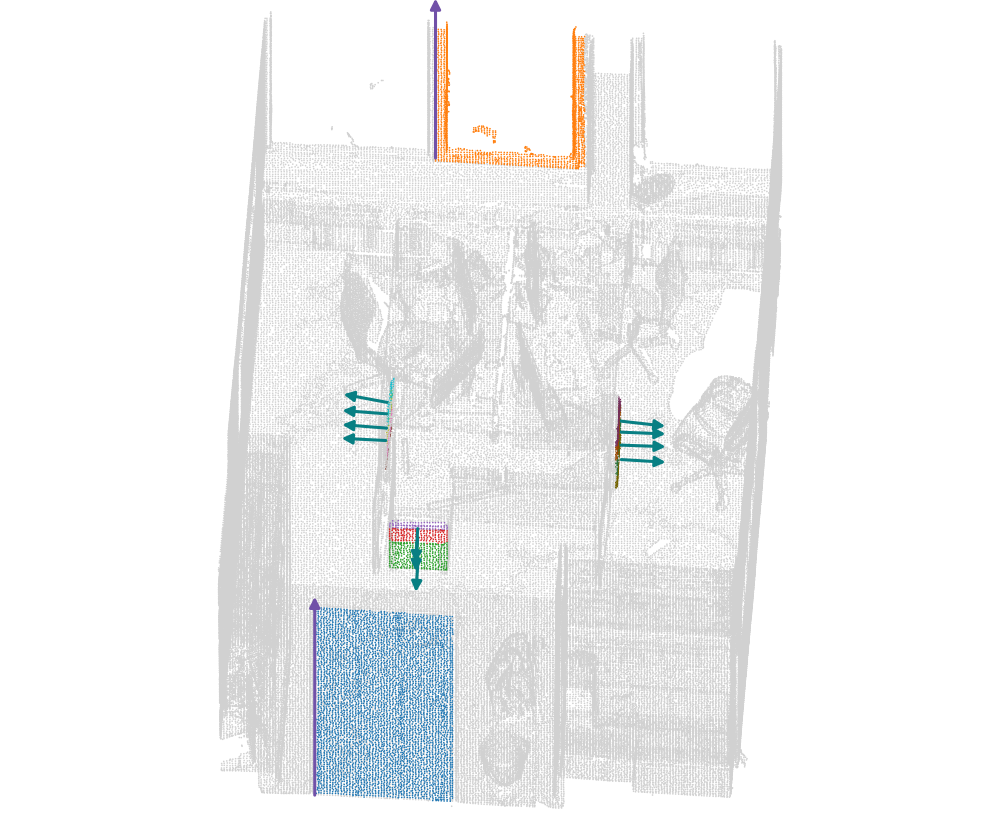} &
\includegraphics[width=.237\linewidth]{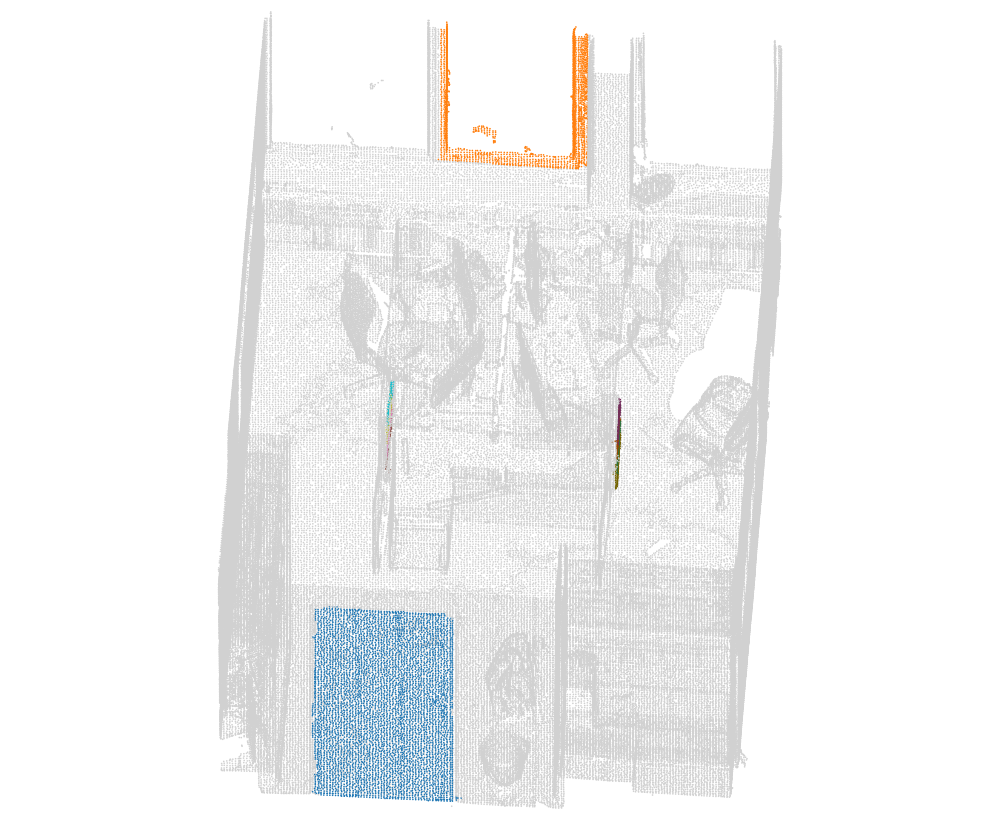} &
\includegraphics[width=.237\linewidth]{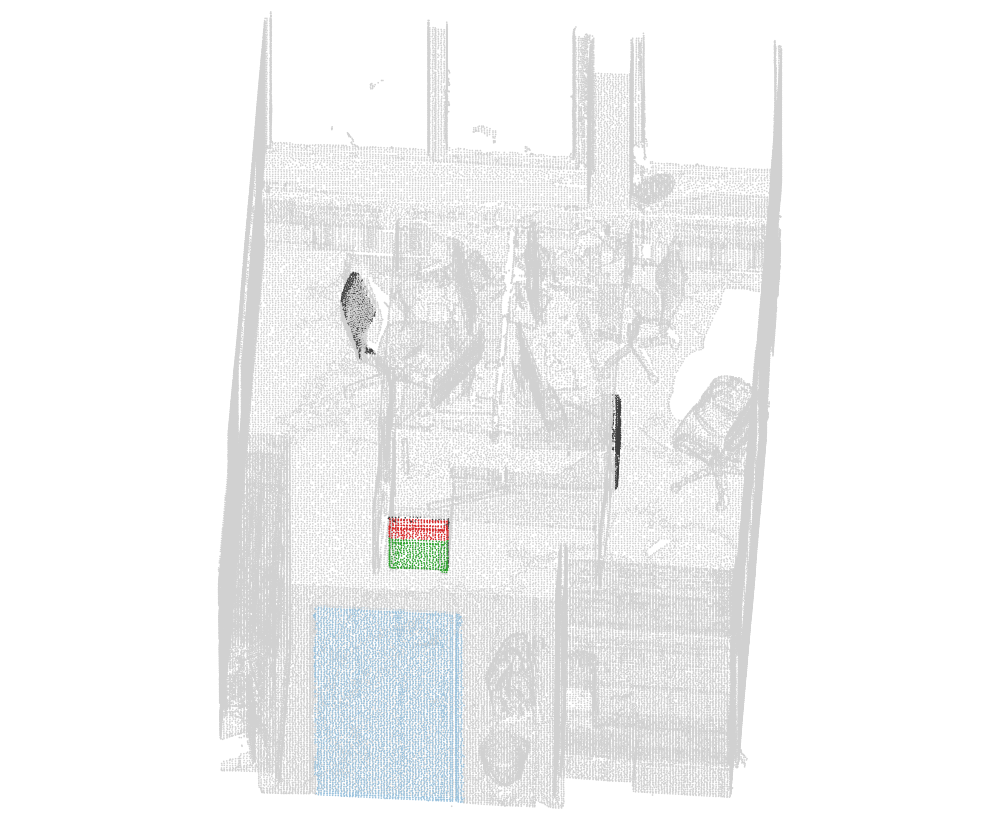} \\
\multicolumn{4}{c}{\scriptsize 419cbe7c11: 13 GT parts; ours 10 matched of 10 shown; REACT3D 2 matched of 6}
\end{tabular}
\caption{\textbf{Comparing part geometry and motion.} Columns show the RGB scene, ground-truth parts and axes, \method, and REACT3D. Purple/teal axes denote rotation/translation. Saturated ground-truth colors mark correct masks and motion; pale colors pass IoU $>0.5$ but fail a motion gate; gray predictions are unmatched. Missing colors reveal missed parts. These favorable scenes are selected by the displayed true-positive advantage; our display threshold is $s\geq0.50$, and all REACT3D predictions are shown.}
\label{fig:react3d-main}
\end{figure}

%% file: tables/react3d_cost.tex
\begin{table}[H]
\centering
\begin{threeparttable}
\caption{\textbf{Inference cost for articulated-part output.} Each system is measured from its declared input through one complete pass per scene, including model loading. Host memory reports the as-shipped peak; GPU logging has unequal coverage and does not support a matched memory ranking. \Cref{app:external} specifies the measurement boundaries and REACT3D's memory refit.}
\label{tab:react3d-cost}
\small
\setlength{\tabcolsep}{7pt}
\begin{tabularx}{\linewidth}{@{}Xrr@{}}
\toprule
\tablehead
\textbf{Measurement} & \textbf{\method} & \textbf{REACT3D} \\
\midrule
Median time (s/scene) & 33.78 & 3,638.1 \\
As-shipped peak host memory (GiB) & 4.70 & 22.94 \\
Logged GPU peak (MiB) & 10,262 & 8,780 \\
Learned models & 2 & 5 \\
Parameters (B) & 0.201 & 5.398 \\
\bottomrule
\end{tabularx}
\end{threeparttable}
\end{table}

%% file: tables/coupling.tex
\begin{table}[H]
\centering
\begin{threeparttable}
\caption{\textbf{Two directed transfers on validation.} Scores are AP (\%); $\Delta$ is the change in percentage points, computed before rounding. Both class-correction rows start from the same proposal union, so their gains are not additive.}
\label{tab:coupling}
\small
\setlength{\tabcolsep}{4pt}
\begin{tabularx}{\linewidth}{@{}Xrrr@{}}
\toprule
\tablehead
\textbf{Controlled change} & \textbf{Before} & \textbf{After} & \textbf{$\Delta$} \\
\midrule
\multicolumn{4}{@{}l}{\emph{Handles $\rightarrow$ parts: motion-gated AP}} \\
Centroid $\rightarrow$ handle-guided origin & 13.74 & \textbf{40.98} & +27.25 \\
\midrule
\multicolumn{4}{@{}l}{\emph{Parts $\rightarrow$ handles: handle AP$_{50}$}} \\
Dense field $\rightarrow$ append associated children & 24.63 & 29.65 & +5.01 \\
Proposal union $\rightarrow$ parts-only class correction & 29.65 & 30.63 & +0.98 \\
\rowours Proposal union $\rightarrow$ full contextual correction & 29.65 & \textbf{30.99} & +1.34 \\
\bottomrule
\end{tabularx}
\begin{tablenotes}[flushleft]\footnotesize
\item The origin intervention fixes masks, scores, classes, and axes (part AP$_{50}=47.93$, axis-gated AP$=43.75$). Class correction fixes child masks and scores and preserves dense detections. The full rule adds a dense-label fallback to part context.
\end{tablenotes}
\end{threeparttable}
\end{table}

%% file: tables/validation_ablations.tex
\begin{table}[H]
\centering
\begin{threeparttable}
\caption{\textbf{Matched learned motion decoders on frozen part inputs.} Learned rows report minimum--maximum \apmotion\ scores (\%) over three training seeds and receive the same predicted handle evidence when marked. All rows retain \ap$=47.93$ because masks are fixed. The bracket in the note gives lower and upper bounds of a paired 95\% scene-jackknife confidence interval for the closest learned head minus the rule (pp); it is distinct from the three-seed score ranges.}
\label{tab:motion-alternatives}
\setlength{\tabcolsep}{5pt}
\begin{tabularx}{\linewidth}{@{}Xcr@{}}
\toprule
\tablehead
\textbf{Motion decoder} & \textbf{Handle cue} & \textbf{\apmotion\ (three-seed range)} \\
\midrule
Reference training-free rule & \checkmark & \textbf{40.98} \\
Box-axis rule with recomputed origin & \checkmark & 41.20 \\
Continuous regression & -- & 17.24--18.24 \\
Continuous regression & \checkmark & 29.61--31.30 \\
Learned candidate selection & -- & 25.38--28.32 \\
Learned candidate selection & \checkmark & 37.27--38.63 \\
Learned selection, box axes only & -- & 25.07--27.84 \\
Learned selection, box axes only & \checkmark & 37.11--38.78 \\
Rule axis + learned edge & -- & 25.36--28.75 \\
Rule axis + learned edge & \checkmark & 37.69--39.36 \\
\bottomrule
\end{tabularx}
\begin{tablenotes}[flushleft]\footnotesize
\item These heads consume frozen query features and geometry, not jointly trained backbones. Discrete learned selection exceeds continuous regression with identical handle evidence, but neither establishes an improvement over the training-free rule. The box-axis variant changes the axis and recomputes the origin; it is not an axis-only control. The closest learned row has paired jackknife 95\% interval $[-4.31,+1.05]$ pp relative to the rule.
\end{tablenotes}
\end{threeparttable}
\end{table}

%% file: tables/mechanisms.tex
\begin{table}[H]
\centering
\begin{threeparttable}
\caption{\textbf{Component changes measured from both ends of the part-motion pipeline.} ``Add gain'' applies one component to the naive pipeline; ``removal loss'' is full decoder minus the result after removing that component. Their difference is a two-end non-additivity diagnostic, not a unique causal interaction. Values are changes in \apmotion\ (pp). The final row's bracket gives the lower and upper bounds of the paired 95\% scene-jackknife confidence interval for the full-minus-naive change, also in pp.}
\label{tab:geometry-controls}
\setlength{\tabcolsep}{5pt}
\begin{tabularx}{\linewidth}{@{}Xrrr@{}}
\toprule
\tablehead
\textbf{Component} & \textbf{Add gain} & \textbf{Removal loss} & \textbf{Two-end difference} \\
\midrule
Per-query argmax selection & $+0.13$ & $+0.19$ & $+0.06$ \\
Largest-CC support cleanup & $-0.11$ & $+2.95$ & $+3.07$ \\
Dense-handle origins & $+23.79$ & $+27.25$ & $+3.45$ \\
Connectivity rescoring & $+1.49$ & $+1.60$ & $+0.11$ \\
\midrule
Full versus naive pipeline & \multicolumn{3}{r@{}}{\textbf{$+28.74$ pp; 95\% CI $[+22.12,+35.37]$}} \\
\bottomrule
\end{tabularx}
\begin{tablenotes}[flushleft]\footnotesize
\item The naive and reference \apmotion\ values are 12.24 and 40.98. Cleanup and handle origins leave mask AP exactly unchanged, as required by their fixed-mask implementation. Cleanup is slightly negative alone but useful in the full decoder, demonstrating non-additivity rather than a contradictory result.
\end{tablenotes}
\end{threeparttable}
\end{table}

%% file: tables/semantic.tex
\begin{table}[H]
\centering
\begin{threeparttable}
\caption{\textbf{Complementary handle localization.} Each configuration appends a child source below the same dense-handle detections without label correction. The corrupted-mask controls preserve child count, mask size and confidence while disrupting localization. Scores are validation \ap\ percentages.}
\label{tab:handle-complementarity}
\setlength{\tabcolsep}{7pt}
\begin{tabularx}{\linewidth}{@{}Xr@{}}
\toprule
\tablehead
\textbf{Proposal source} & \textbf{Handle \ap} \\
\midrule
Dense detections only & 24.63 \\
Dense + spatially permuted child masks & 24.63 \\
Dense + random size-matched child masks & 24.63 \\
\rowours Dense + query-associated child proposals & \textbf{29.65} \\
\bottomrule
\end{tabularx}
\begin{tablenotes}[flushleft]\footnotesize
\item The real child union retains all 140 GT handles matched by dense incumbents and adds 53 matches, while preserving the incumbent TP/FP ranking prefix. Its gain demonstrates complementary localization, not an isolated effect of parent conditioning; \Cref{tab:app-association} separates conditioning from the readout used to associate children.
\end{tablenotes}
\end{threeparttable}
\end{table}

%% file: tables/contextual_labels.tex
\begin{table}[H]
\centering
\begin{threeparttable}
\caption{\textbf{Contextual child-label controls.} Child masks and scores are fixed; only labels of appended children can change. Scores are validation \ap\ percentages.}
\label{tab:context-controls}
\setlength{\tabcolsep}{5pt}
\begin{tabularx}{\linewidth}{@{}Xccrr@{}}
\toprule
\tablehead
\textbf{Label rule} & \textbf{Part cue} & \textbf{Dense cue} & \textbf{\ap} & \textbf{$\Delta$} \\
\midrule
No correction & -- & -- & 29.65 & -- \\
Parts only & \checkmark & -- & 30.63 & $+0.98$ \\
Dense incumbents only & -- & \checkmark & 30.66 & $+1.01$ \\
\rowours Full contextual rule & \checkmark & \checkmark & \textbf{30.99} & \textbf{$+1.34$} \\
Fallback only when no part speaks & \checkmark & \checkmark & 30.99 & $+1.34$ \\
Part classes permuted within scene & corrupted & \checkmark & 28.62 & $-1.03$ \\
Part classes randomized & corrupted & \checkmark & 28.90 & $-0.75$ \\
\bottomrule
\end{tabularx}
\begin{tablenotes}[flushleft]\footnotesize
\item The two cues are complementary but overlap: their individual gains must not be summed. Restricting fallback changes one child and is AP-identical on this split. In the last two rows, masks and scores are frozen and only the part-class channel is corrupted; dense labels remain intact. They relabel more children than the reference rule, but lower AP below the no-correction union, so relabel count alone is not evidence of useful context. Dense incumbents themselves are never relabelled.
\end{tablenotes}
\end{threeparttable}
\end{table}

%% file: tables/robustness.tex
\begin{table}[H]
\centering
\begin{threeparttable}
\caption{\textbf{Observed variation of the directed effects.} All values are AP changes in pp. ``Seed span'' is the maximum minus minimum paired gain across $n$ training draws, not a confidence interval. Brackets give the lower and upper bounds of the paired 95\% scene-jackknife confidence interval (CI) for the reference gain. A CI containing zero leaves the direction of change unresolved.}
\label{tab:coupling-robustness}
\setlength{\tabcolsep}{4pt}
\begin{tabularx}{\linewidth}{@{}>{\raggedright\arraybackslash}Xrrl@{}}
\toprule
\tablehead
\textbf{Mechanism} & \textbf{Reference gain} & \textbf{Seed span} & \textbf{95\% scene CI} \\
\midrule
Dense handles $\rightarrow$ hinge origin & $+27.25$ & 2.11 ($n=4$) & $[+21.86,+32.63]$ \\
Child proposals appended to dense handles & $+5.01$ & 1.08 ($n=3$) & $[+1.47,+8.56]$ \\
Contextual child-label vote & $+1.34$ & 1.15 ($n=3$) & $[+0.37,+2.31]$ \\
\bottomrule
\end{tabularx}
\begin{tablenotes}[flushleft]\footnotesize
\item The child family varies the part-associated child predictor with dense incumbents fixed. Its append increments are 5.01/4.89/3.94 pp; its vote increments are 1.34/0.19/0.38 pp. The append interval compares the reference child union against dense detections alone; the gain remains positive in every leave-one-scene-out evaluation. This supports a complementary proposal source, not an isolated conditioning effect.
\end{tablenotes}
\end{threeparttable}
\end{table}

%% file: figures/qualitative_hinge.tex
\begin{figure}[t]
\centering
\begin{tikzpicture}[inner sep=0pt,outer sep=0pt]
\node[inner sep=0pt,outer sep=0pt] (image) {\includegraphics[width=\linewidth,trim=0bp 30bp 80bp 72bp,clip]{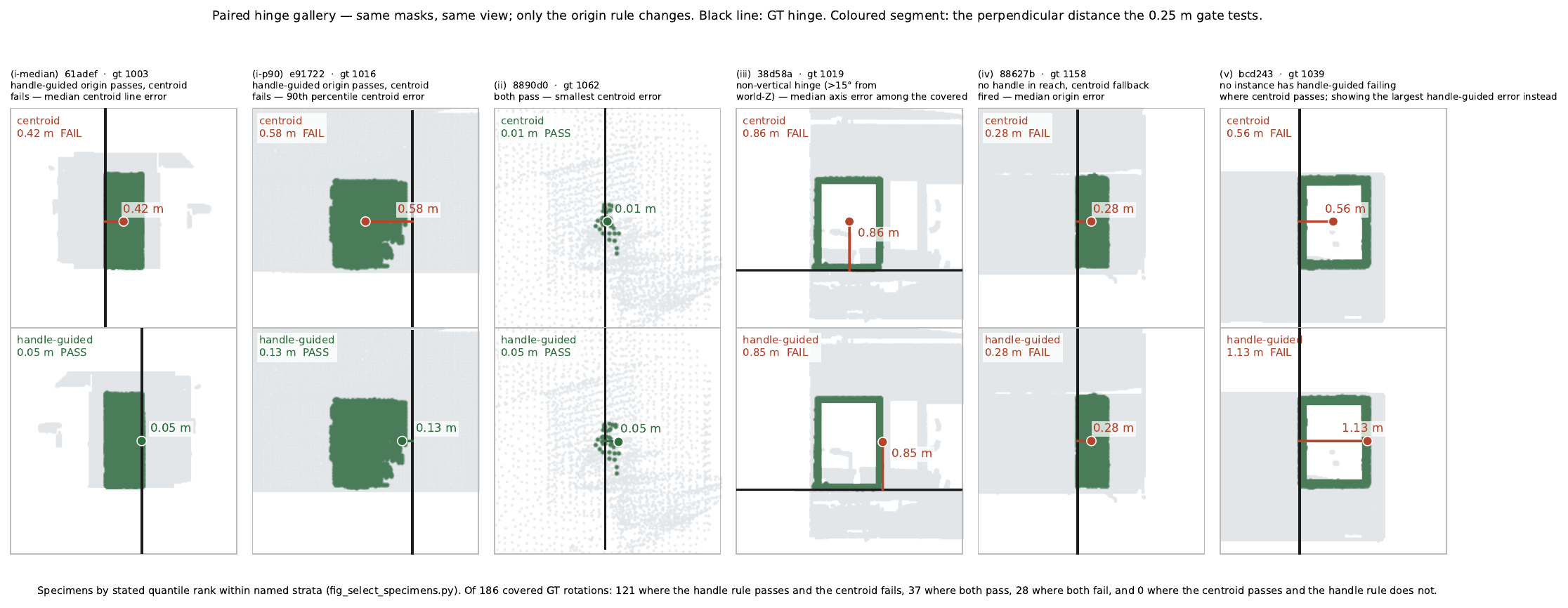}};
\begin{scope}[shift={(image.south west)},x={($(image.south east)-(image.south west)$)},y={($(image.north west)-(image.south west)$)}]
\foreach \x/\t in {0/{Median centroid\\error},.1667/{90th-percentile\\centroid error},.3334/{Both rules\\pass},.5001/{Non-vertical\\hinge},.6668/{Missing-handle\\fallback},.8333/{Largest handle-\\guided error}} {
  \node[anchor=south,font=\sffamily\fontsize{6}{7}\selectfont,align=center,text width=.158\linewidth,text=black!75] at (\x+.0833,1.025) {\t};
}
\end{scope}
\end{tikzpicture}
\caption{\textbf{Paired hinge-origin cases on validation scenes.} Each column compares centroid and handle-guided origins for the same predicted part. Black denotes the ground-truth hinge, and the colored segment depicts the perpendicular origin error. The first two columns are centroid-error ranks 61 and 109 among 121 rotations for which handle guidance passes but the centroid fails (median and 90th percentile). The remaining selections are the smallest centroid error among 37 both-pass cases, median axis error among 12 non-vertical hinges, median origin error among 19 missing-handle fallbacks, and the largest handle-guided error among 186 covered rotations. No case has a passing centroid but failing handle-guided origin in this population; the last column retains a large residual failure instead.}
\label{fig:qual-hinge}
\end{figure}

%% file: figures/qualitative_handles.tex
\begin{figure}[t]
\centering
\begin{tikzpicture}[inner sep=0pt,outer sep=0pt]
\node[inner sep=0pt,outer sep=0pt] (image) {\includegraphics[width=\linewidth,trim=110bp 54bp 113bp 73bp,clip]{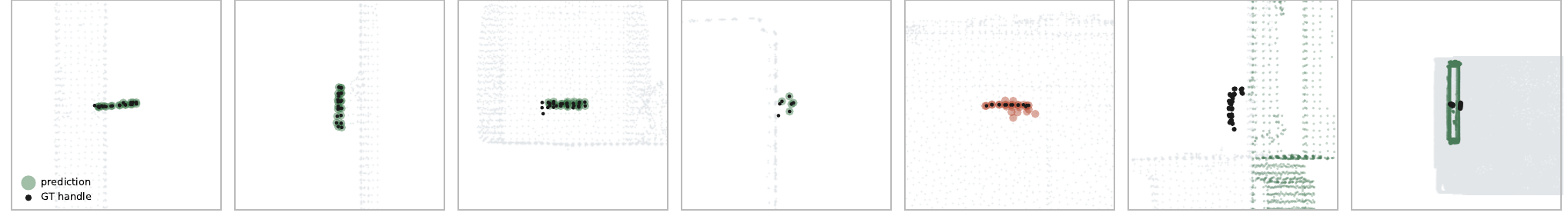}};
\begin{scope}[shift={(image.south west)},x={($(image.south east)-(image.south west)$)},y={($(image.north west)-(image.south west)$)}]
\foreach \x/\t in {0/{Dense success\\median size},.1429/{Dense success\\above median},.2857/{Child-only\\median size},.4286/{Child-only\\smallest size},.5714/{Over-extended\\dense mask},.7143/{Contextual\\label update},.8571/{Unmatched\\relabeled child}} {
  \node[anchor=south,font=\sffamily\fontsize{6}{7}\selectfont,align=center,text width=.135\linewidth,text=black!75] at (\x+.0714,1.06) {\t};
}
\end{scope}
\end{tikzpicture}
\caption{\textbf{Dense and child handle complementarity on validation scenes.} The selected strata show dense successes, handles recovered only after appending child proposals, an over-extended dense prediction, and contextual child-label updates. The final panel is deliberately a relabeled child without an IoU$>0.5$ ground-truth match, making explicit that label accuracy is only directly assessable for the matched subset. Specimens are deterministic quantile selections within their stated strata.}
\label{fig:qual-handles}
\end{figure}

%% file: tables/leaderboard.tex
\begin{table}[H]
\centering
\begin{threeparttable}
\caption{\textbf{Challenge leaderboard: movable-part motion result.} All 11 public team entries and all four published metrics. Scores are percentages; ranking is by \mao.}
\label{tab:leaderboard-mov}
\small
\setlength{\tabcolsep}{4.6pt}
\begin{tabularx}{\linewidth}{@{}c>{\raggedright\arraybackslash}p{3.8cm}*{4}{>{\raggedleft\arraybackslash}X}@{}}
\toprule
\tablehead
\textbf{Rank} & \textbf{Team} & \ap & \apaxis & \aporigin & \mao \\
\midrule
\rowours \textbf{1} & \textbf{TnG (ours)} & \textbf{55.38} & \textbf{52.34} & \textbf{49.76} & \textbf{48.28} \\
2 & coreMany & 46.68 & 43.15 & 43.28 & 40.56 \\
3 & teamzb & 47.20 & 44.99 & 38.55 & 37.11 \\
4 & core & 44.44 & 41.05 & 39.55 & 36.89 \\
5 & ZZZ & 47.72 & 44.81 & 37.77 & 35.72 \\
6 & linxii & 43.52 & 41.35 & 35.47 & 33.60 \\
7 & coreN & 43.70 & 40.60 & 35.94 & 33.38 \\
8 & ZeroR & 39.53 & 37.25 & 30.14 & 28.61 \\
9 & wwwjh & 43.66 & 37.98 & 31.54 & 26.34 \\
10 & ShinNam! & 35.52 & 33.73 & 29.04 & 24.67 \\
11 & ttt & 30.58 & 28.38 & 25.99 & 24.00 \\
\bottomrule
\end{tabularx}
\begin{tablenotes}[flushleft]\footnotesize
\item Source: \href{https://art3d-challenge.mooo.com/web/challenges/challenge-page/1/leaderboard/}{official Articulate3D leaderboard}~\citep{articulate3dleaderboard2026}, accessed September 6, 2026. Superscripts A, O, and AO denote axis, origin, and joint motion checks, respectively.
\end{tablenotes}
\end{threeparttable}
\end{table}

\begin{table}[H]
\centering
\begin{threeparttable}
\caption{\textbf{Challenge leaderboard: handle-detection result.} All eight public team entries. The board provides \ap\ only; scores are percentages.}
\label{tab:leaderboard-handle}
\small
\setlength{\tabcolsep}{8pt}
\begin{tabularx}{0.60\linewidth}{@{}cXr@{}}
\toprule
\tablehead
\textbf{Rank} & \textbf{Team} & \ap \\
\midrule
\rowours \textbf{1} & \textbf{TnG (ours)} & \textbf{34.46} \\
2 & coreMany & 32.86 \\
3 & core & 32.60 \\
4 & teamzb & 31.10 \\
5 & ZZZ & 30.69 \\
6 & coreN & 28.86 \\
7 & ZeroR & 25.71 \\
8 & moon & 24.80 \\
\bottomrule
\end{tabularx}
\begin{tablenotes}[flushleft]\footnotesize
\item Same source and access date as \Cref{tab:leaderboard-mov}. The public movable-part-motion and handle-detection boards contain different participant sets.
\end{tablenotes}
\end{threeparttable}
\end{table}

%% file: sections/05_discussion.tex
\section{Discussion and limitations}
\label{sec:discussion}

\paragraph{What coupling contributes}
Handles and parts provide different forms of evidence: handle locations constrain a part's attachment side, while part-associated proposals and motion classes improve handle detection and interpretation. The largest controlled gain comes from hinge-side evidence rather than segmentation: changing origins alone raises motion AP by 27.25 points (\Cref{tab:coupling,tab:app-origin-control}). The reverse transfer separates complementary localization from contextual class correction. No iterative refinement is required because the two directions read and update different attributes. Within the geometric decoder, however, support fitting and origin selection interact; their component gains are not additive (\Cref{tab:geometry-controls}).

\paragraph{Physical and perceptual limits}
The decoder assumes approximately planar parts and upright rotational axes. Horizontal hinges, tilted or curved mechanisms, missing handles and fragmented support can violate these assumptions; proximity-based association can also select a neighboring handle (\Cref{fig:app-hinges,fig:app-failures}). High hinge-candidate coverage does not resolve missing instances or incorrect motion classes. In a first-failure census of 390 ground-truth parts, 139 lack a matching predicted mask, compared with 14 axis failures and 14 origin failures (\Cref{tab:app-coverage}). Very small handles remain difficult, and changing a handle's class cannot repair its location or extent. These observations motivate better instance coverage and motion hypotheses beyond upright planar mechanisms.

\paragraph{Scope of the measured benefits}
The proposal gain demonstrates complementary localization, not a separately established advantage of parent conditioning (\Cref{tab:app-association}). Label correction improves the reference configuration, but its magnitude changes across child-training draws (\Cref{tab:coupling-robustness}). Learned-decoder controls show that handle evidence helps several decoder families; overlapping paired intervals do not establish a statistically resolved advantage for the fixed geometric rule over the strongest selection heads (\Cref{tab:motion-alternatives,tab:app-learned-two-bases}). We evaluate on one development dataset, and validation-informed model selection limits broader generalization claims. The REACT3D comparison evaluates a common output on the same scenes, while retaining the systems' different inputs and training conditions (\Cref{sec:external,app:external}).

\section{Conclusion}
\label{sec:conclusion}

We presented \method for recovering movable parts, motion and handles as a joint description of interaction in static 3D scenes. Predicted handles guide training-free hinge selection; part-associated proposals expand handle coverage, and part context refines their motion labels. Fixed-input comparisons, repeated training, complete-scene comparisons and paired visualizations characterize the benefits and limits of these directed transfers. Together with the first-place Articulate3D Challenge result, the evidence supports geometric and semantic coupling as an effective strategy for 3D interaction understanding. Further progress requires improved instance coverage and motion models beyond upright planar mechanisms.

%% file: sections/06_appendix.tex
\section{Implementation and Evaluation Details}
\label{sec:appendix-implementation}
\label{app:implementation}

This appendix documents the reference validation configuration, the controls behind the reported coupling effects, and additional qualitative evidence. Unless stated otherwise, all results use the 42-scene public Articulate3D validation split. They are development-set evidence rather than an untouched generalization estimate.

\subsection{Independent perception models}

\method uses three independently trained networks with the same RGB point cloud and surface normals as input. Each uses a Volt-B voxel Transformer initialized from ScanNet++ pretraining~\citep{yilmaz2026volt}, with 2\,cm voxels and $5\!\times\!5\!\times\!5$ voxel patches. The \emph{standalone part predictor} uses an SPFormer decoder~\citep{sun2023spformer} over graph-partitioned superpoints~\citep{felzenszwalb2004efficient}. Its 200 queries predict a movable-part mask, rotation/translation class, and confidence; Hungarian matching uses classification, binary cross-entropy, Dice, and score terms. These predictions, $\mathcal P$, provide every final movable-part instance.

The \emph{dense handle predictor} emits a pointwise background/rotation-handle/translation-handle field. It is trained with class-weighted cross-entropy, label smoothing, and multiclass Lov\'asz loss~\citep{berman2018lovasz}. At inference, class-wise radius-graph components (2.5\,cm radius, at least three points) become handle instances, scored by mean assigned-class probability. These dense instances $\mathcal H_d$ remain in the final handle set and provide the geometric cue for motion decoding.

The third, \emph{joint part-handle predictor}, decodes its own parent parts $\mathcal Q$ and a fine child-handle mask for each retained parent query; it does not replace $\mathcal P$. A parent head and child head read the same query features, and the child head does not consume a thresholded parent mask. For query $\boldsymbol q_j$ and voxel feature $\boldsymbol f_v$, its child probability is
\begin{equation}
 p_{jv}=\operatorname{sigmoid}\!\left(g_q(\boldsymbol q_j)^\mathsf{T}g_v(\boldsymbol f_v)\right),
\end{equation}
where $g_q$ and $g_v$ are learned projections to a common feature space. Parents alone receive Hungarian assignment. Child supervision uses binary cross-entropy, Dice, and twice a Tversky loss with false-positive and false-negative weights 0.7 and 0.3~\citep{salehi2017tversky}; a voxel is positive if it contains any target-child point. This retains small supports that majority pooling could remove.

\subsection{Deterministic coupling}

For each standalone part mask $M_i$, we retain its largest 5\,cm-radius connected component $S_i$ only for fitting, and rescale its confidence $s_i$ to $\bar{s}_i=s_i|S_i|/|M_i|$. The output mask and class remain unchanged. Fitting uses at most 20,000 support points, with deterministic subsampling above that cap. Their mean is $\boldsymbol\mu_i$. A PCA plane and in-plane minimum-area rectangle supply orientation and extents; the anchor remains $\boldsymbol\mu_i$, rather than the rectangle centre. A rotational part uses the vertical unit vector $\boldsymbol e_z=(0,0,1)^\top$; a translating part uses the box's thinnest direction. The decoder contains no learned motion regressor, although its masks and handle locations are learned.

For a rotational part, the nearest dense-handle centroid is accepted within 0.5\,m of $S_i$. Four candidate lines run parallel to the predicted axis through anchors obtained by adding signed half-extents to $\boldsymbol\mu_i$ along the two least-aligned box directions. The selected line is farthest from the handle in perpendicular distance, and the emitted origin is the projection of $\boldsymbol\mu_i$ onto that line. Without an accepted handle, the origin is $\boldsymbol\mu_i$; translations use the same mean. For numerical stability in the evaluator, emitted axes receive a deterministic $10^{-6}$-radian perturbation and are renormalized after geometry decoding. The construction assumes approximately planar surfaces and upright rotational axes; it does not model arbitrary mechanisms.

The joint branch retains child voxels above probability 0.30 and retrieves each child by its source-query index after parent filtering. Children are appended, not mask-merged, to dense handles with their parent class and a score equal to 0.05 times parent confidence. Only an appended child's class can change. Eligible standalone parts contain at least 90\% of the child and have calibrated confidence at least 0.3. The highest-confidence eligible part assigns its class; only when no part qualifies does the dense handle with largest containment, at least 90\%, provide a fallback class. Otherwise the initial child label remains. Masks and scores remain fixed, and dense detections are never relabelled. Dense-handle centroids guide part origins once; joint-model parents supply child proposals, and standalone parts supply child-label context. Final handles are not fed back into motion decoding.

\subsection{Optimization, selection, and metrics}

All three networks are optimized separately with AdamW~\citep{loshchilov2019adamw} for 400 epochs, peak learning rate $3\times10^{-4}$, cosine one-cycle scheduling, batch size two with eight-step gradient accumulation, and EMA decay 0.999. Weight decay is 0.1 for the part and joint models and 0.05 for dense handles. We use spatial crops, rotations, scaling, and color augmentation; the dense model additionally uses flips, jitter, and elastic distortion. Its foreground targets are dilated by 0.10\,m for the first 50\% of training, 0.04\,m for the next 30\%, and not dilated for the final 20\%. This is a configuration choice, not a claimed contribution.

The part model is selected by validation motion-gated AP at 70\% of training; the dense branch uses a segmentation-selected checkpoint and the joint branch its final weights. We report macro AP at IoU 0.5. Axis-gated AP additionally requires sign-invariant axis error below $15^\circ$. For \apmotion, a rotation must pass two origin checks: the predicted origin's distance to the annotated axis line and the annotated origin's distance to the predicted axis line must both be below 0.25\,m. Translations have no origin clause. All axes are normalized before these projections. The evaluator also exposes a stricter Euclidean-origin variant, which checks full representative-origin displacement. The reference system scores 38.99 under that variant, versus 40.98 with line distance. These conventions are not interchangeable; all coupling ablations use the latter.

\section{Additional Quantitative Analysis}
\label{sec:appendix-quantitative}
\label{app:controls}

\subsection{Controls and uncertainty}
\label{sec:appendix-sources}

The main report presents reference gains and paired scene uncertainty in \Cref{tab:coupling-robustness}, with localization and class-evidence controls in \Cref{tab:handle-complementarity,tab:context-controls}. Here we resolve the effects by motion class and vary the perception models that supply each cue.

\input{tables/coupling_per_class}

\Cref{tab:coupling-per-class} shows that handle guidance improves rotational motion AP by 54.49 pp while leaving translations unchanged, as expected from an origin-only intervention. Conversely, contextual correction improves translation-handle AP by 2.71 pp; the rotation-handle difference is $-0.02$ pp and its paired interval includes zero. The class breakdown explains the macro gains without implying that every output class benefits equally. Brackets describe scene-sampling uncertainty in AP differences, not training-seed variability.

\input{tables/app_source_stability}

\Cref{tab:app-source-variation} changes the target predictor while fixing its complementary cue. Handle guidance improves the reference, joint-head, and random-initialization part configurations; the same child source improves both reference and weak dense-handle configurations. Random-initialization rows also use a different training schedule, so they are source-configuration comparisons, not isolated tests of pretraining. \Cref{tab:app-dense-repeats} varies the dense predictor while fixing the part and child sources. Proposal augmentation and contextual correction improve handle AP in every row, although dense-handle AP and part-motion AP need not order the rows identically because hinge selection uses local handle position rather than the whole mask-and-class ranking. The child-draw values in \Cref{tab:coupling-robustness} are not pooled with these repeats.

\subsection{Information flow and child association}
\label{sec:appendix-directions}

\input{tables/coupling_matrix}

\Cref{tab:coupling-matrix} holds the proposal union fixed and switches the two directed transfers independently. Handle guidance changes only motion AP, and contextual correction changes only handle AP. Their measured interaction is zero because decoded origins are not used for label correction and corrected handles are not fed back into decoding. This confirms the implemented information flow rather than an iterative refinement effect.

\input{tables/app_association}

The main-text coupling control verifies output ownership: dense-handle guidance affects part origins, whereas contextual labels affect handles. \Cref{tab:app-association} separates parent conditioning from association. Under matched shared-field readout, the conditioning contrast changes sign after correction and both paired intervals include zero. The alternate-dense association control instead changes which children are emitted and ranked while holding dense instances fixed; its 1.24-pp union difference concerns association and ranking, not newly learned mask geometry. These results support the implemented proposal source without isolating conditioning alone as its cause.

\subsection{Motion-decoder controls}
\label{sec:appendix-decoders}

\input{tables/app_decoder_controls}

\Cref{tab:app-origin-control} isolates handle guidance by preserving masks, classes, scores, and axes; equal mask and axis AP confirm that only the origin rule changes. This complements the component interventions in \Cref{tab:geometry-controls}, which also alter fitting support and instance ranking. The learned-decoder controls below extend the three-seed ranges in \Cref{tab:motion-alternatives} with paired comparisons on a second part-model draw.

\paragraph{Learned-decoder protocol}
The learned heads use frozen part-query features, the fitted box frame, and gravity-alignment features; handle-aware variants additionally receive the same predicted-handle offset used by the rule. They train on predictions from the 195 training scenes and are evaluated on the 42 validation scenes without changing masks, classes, or scores. Three seeds vary initialization and a 20\% held-out training-scene fold. Within a seed, the checkpoint minimizes loss on that fold; the displayed best-seed summaries therefore report the highest validation motion AP among the three seeds and are descriptive, not independently selected estimates. Optimization uses AdamW for 1,500 full-batch epochs with cosine scheduling, learning rate $10^{-3}$, and weight decay $10^{-4}$. Continuous regression predicts axis and origin; selection chooses axis and edge candidates; the fixed-axis variant learns only edge choice. This evaluates decoding from a fixed representation, not end-to-end retraining.

\Cref{tab:app-learned-two-bases} compares matched learned heads with the training-free rule on two fixed part-model draws. Learned alternatives have lower observed point estimates, but every displayed paired interval includes zero. The evidence supports a useful training-free decoder in this setting, rather than a statistically resolved or general advantage over learned decoders.

\subsection{Coverage and failure accounting}
\label{sec:appendix-coverage}
\label{app:coverage}

\input{tables/app_coverage}

Candidate availability and candidate selection are separate diagnostics. The matched-rotation population can contain several predictions for one ground-truth instance, whereas the GT-mask diagnostic uses one mask per annotated rotation. Thus, the high candidate-coverage fractions in \Cref{tab:app-coverage} are not macro-AP ceilings. Of all 390 ground-truth parts, 64.4\% have a matching mask and 57.2\% pass the joint motion criterion. The first-failure census shows where end-to-end instances are lost: missing movable-part masks dominate, especially for translations. The joint union covers 193 of 388 ground-truth handles; its smallest size quartile contains at most 11 points and has the lowest coverage. These are descriptive populations, not additive oracle gains.

Uncovered rotations have median best-mask IoU 0.30, versus 0.10 for translations. All 12 rotation-axis failures involve non-vertical hinges, and 12 of 14 origin failures lie between 0.26 and 0.41\,m, close to the 0.25\,m gate. Of 75 smallest-quartile handle misses, 38 have no overlapping prediction. Conversely, 37 of 96 largest-quartile handles have an overlapping prediction at least twice their size. Improving hinge placement alone therefore cannot remove the coverage bottleneck.

\FloatBarrier
\section{Additional Qualitative Analysis}
\label{sec:appendix-visual}
\label{app:qualitative}

\subsection{Coverage and geometric components}

\Cref{fig:appendix-coverage-selection} visualizes the census in \Cref{tab:app-coverage}: missing instances limit the gains available to motion decoding, and small handles remain difficult. \Cref{fig:qual-baseline} compares the naive and full configurations on common GT instances, including a case where the naive origin is closer and a newly covered instance whose origin still fails. Predicted instance sets can differ between these configurations.

\input{figures/final_coverage_selection}
\input{figures/qualitative_baseline}

\FloatBarrier
\subsection{Weaker prediction sources and representative failures}

\Cref{fig:appendix-weak-handle-field} fixes the scene crop and compares dense-handle fields from the reference and random-initialization configurations. The expanded galleries below show the corresponding changes in part source, failures of the upright-axis prior, origin-selection errors, and small or over-extended handle predictions. These deterministic selections explain mechanisms rather than estimate their prevalence.

\input{figures/final_weak_handle_field}

\FloatBarrier
\subsection{Expanded galleries}

\Cref{fig:app-hinges} enlarges the paired hinge examples, separating missing handle evidence from violations of the upright-axis assumption. \Cref{fig:app-components} connects the component study in \Cref{tab:geometry-controls} to its geometric action: detached support can distort fitted candidates, while the median-change example is unaffected. A well-fitted surface alone still leaves the centroid-origin ambiguity.

\Cref{fig:app-failures} visualizes the census categories in \Cref{tab:app-coverage}. Missing masks cannot be repaired by motion decoding, and non-vertical hinges cannot be corrected by selecting another vertical line. \Cref{fig:app-weaker-parts} changes the part source while holding dense handles fixed, illustrating the configuration comparisons in \Cref{tab:app-source-variation}. Each within-row centroid-versus-guidance pair uses the same predicted mask; different networks need not predict identical masks.

\Cref{fig:app-handles} enlarges dense successes, child-only recoveries, and class-correction examples. Its final unmatched child shows why correctness among matched, relabelled handles cannot be extended to every corrected proposal. Each gallery's caption specifies its selection population; these examples explain mechanisms rather than estimate their prevalence.

\input{figures/appendix_galleries}
\FloatBarrier

%% file: tables/coupling_per_class.tex
\begin{table}[!htbp]
\centering
\begin{threeparttable}
\caption{\textbf{Per-class localization of the directed effects.} Absolute rows are AP (\%); difference rows and brackets are pp. Each bracket is the paired 95\% scene-jackknife interval for the difference immediately above it. Macro AP averages the rotation and translation classes; differences use unrounded scores.}
\label{tab:coupling-per-class}
\small
\setlength{\tabcolsep}{5pt}
\begin{tabularx}{\linewidth}{@{}Xrrr@{}}
\toprule
\tablehead
\textbf{Configuration or difference} & \textbf{Rotation} & \textbf{Translation} & \textbf{Macro} \\
\midrule
Part-motion centroid origin & 1.88 & 25.59 & 13.74 \\
Part-motion dense-handle origin & 56.37 & 25.59 & 40.98 \\
\quad Dense-handle minus centroid & $+54.49$ & $+0.00$ & $+27.25$ \\
\quad 95\% paired CI & $[+43.71,+65.27]$ & $[0.00,0.00]$ & $[+21.86,+32.63]$ \\
\addlinespace
Handle child union & 48.56 & 10.74 & 29.65 \\
Handle contextual correction & 48.54 & 13.45 & 30.99 \\
\quad Correction minus union & $-0.02$ & $+2.71$ & $+1.34$ \\
\quad 95\% paired CI & $[-0.15,+0.10]$ & $[+0.80,+4.62]$ & $[+0.37,+2.31]$ \\
\bottomrule
\end{tabularx}
\begin{tablenotes}[flushleft]\footnotesize
\item Origins are not evaluated for translations, so their unchanged motion output gives an exactly zero interval. The handle-label gain is concentrated on translation handles. Class-resolved intervals describe this validation population, not generalization beyond it.
\end{tablenotes}
\end{threeparttable}
\end{table}

%% file: tables/app_source_stability.tex
\begin{table}[!htbp]
\centering\small
\caption{Coupling under different perception configurations. Scores are validation AP (\%); the complementary cue is fixed within each block. The random-initialization configurations also use longer training schedules, so these rows do not isolate pretraining.}
\label{tab:app-source-variation}
\begin{tabularx}{\linewidth}{@{}Xrr@{}}
\toprule\tablehead
Prediction source & Without cue & With cue \\
\midrule
\multicolumn{3}{@{}l}{\emph{Motion AP: fixed dense handles, varying part predictor}} \\
Standalone reference parts & 13.74 & 40.98 \\
Joint-model part head & 12.26 & 37.33 \\
Random-initialization part configuration & 5.86 & 25.42 \\
\midrule
\multicolumn{3}{@{}l}{\emph{Handle AP: fixed child proposals, varying dense predictor}} \\
Reference dense model & 24.63 & 29.65 \\
Random-initialization dense configuration & 1.75 & 13.63 \\
\bottomrule
\end{tabularx}
\end{table}

\begin{table}[!htbp]
\centering\small
\caption{Repeated dense-handle training with part and child predictions fixed. All scores are AP (\%). The first three columns measure handles; the last measures part motion using that row's dense field as the geometric cue.}
\label{tab:app-dense-repeats}
\setlength{\tabcolsep}{4pt}
\begin{tabularx}{\linewidth}{@{}Xrrrr@{}}
\toprule\tablehead
Dense-model draw & Dense handles & + Children & + Context & Part motion \\
\midrule
\rowours Reference & 24.63 & 29.65 & 30.99 & 40.98 \\
Repeat 1 & 23.76 & 27.68 & 28.81 & 40.66 \\
Repeat 2 & 22.83 & 27.67 & 28.90 & 40.46 \\
Repeat 3 & 25.16 & 29.70 & 30.82 & 39.51 \\
\bottomrule
\end{tabularx}
\end{table}

%% file: tables/coupling_matrix.tex
\begin{table}[H]
\centering
\begin{threeparttable}
\caption{\textbf{Orthogonal one-pass coupling controls.} The child-proposal union is fixed in every cell; the factorial switches only dense-handle guidance for part origins and contextual child-label voting for handles. Scores are validation percentages. Brackets in the note give lower and upper bounds of paired 95\% scene-jackknife confidence intervals for each signal's on-minus-off AP change, in pp.}
\label{tab:coupling-matrix}
\setlength{\tabcolsep}{5pt}
\begin{tabularx}{\linewidth}{@{}Xccrr@{}}
\toprule
\tablehead
& \multicolumn{2}{c}{\textbf{Switched signal}} & \multicolumn{2}{c}{\textbf{Ranking metric}} \\
\cmidrule(lr){2-3}\cmidrule(l){4-5}
\tablehead
\textbf{Configuration} & \textbf{Handle-guided origin} & \textbf{Context vote} & \textbf{Part motion \apmotion} & \textbf{Handle \ap} \\
\midrule
Neither signal & -- & -- & 13.74 & 29.65 \\
Handle-guided hinge only & \checkmark & -- & \textbf{40.98} & 29.65 \\
Contextual vote only & -- & \checkmark & 13.74 & \textbf{30.99} \\
\rowours Both signals (reference) & \checkmark & \checkmark & \textbf{40.98} & \textbf{30.99} \\
\bottomrule
\end{tabularx}
\begin{tablenotes}[flushleft]\footnotesize
\item Handles change part-motion AP by $+27.25$ pp (paired scene jackknife 95\% interval $[+21.86,+32.63]$); voting changes handle AP by $+1.34$ pp ($[+0.37,+2.31]$). The measured interaction is exactly zero on both metrics: handle guidance changes origins only, while voting reads masks and classes. This is a factorial control of two directed transfers, not evidence of iterative refinement.
\end{tablenotes}
\end{threeparttable}
\end{table}

%% file: tables/app_association.tex
\begin{table}[!htbp]
\centering\small
\caption{Conditioning and association are separate comparisons. Scores are handle AP (\%). Panel (a) holds the shared-field readout fixed when changing conditioning. Panel (b) holds a different dense prediction field fixed and changes child association; its absolute scores are not the reference chain.}
\label{tab:app-association}
\setlength{\tabcolsep}{4pt}
\begin{tabularx}{\linewidth}{@{}Xrr@{}}
\toprule\tablehead
Predictor / readout & Proposal union & After correction \\
\midrule
\multicolumn{3}{@{}l}{\emph{(a) Parent conditioning with matched shared-field readout}} \\
Unconditioned child head & 28.07 & 29.78 \\
Conditioned child head & 28.41 & 29.26 \\
\midrule
\multicolumn{3}{@{}l}{\emph{(b) Parent-child association on an alternate dense-model basis}} \\
Output-position association & 27.72 & 29.18 \\
Query-index association & 28.96 & 30.37 \\
\bottomrule
\end{tabularx}
\par\smallskip\begin{minipage}{\linewidth}\footnotesize
In (a), conditioned minus unconditioned is +0.34 pp before correction (95\% CI $[-0.99,+1.66]$) and $-0.52$ pp after ($[-2.66,+1.62]$). In (b), all 443 dense instances are identical. Query and position association emit 4,129 and 3,317 child instances; among 1,691 shared distinct child masks, 1,688 receive different scores. The 1.24-pp union gain reflects association and ranking, not newly learned mask geometry.
\end{minipage}
\end{table}

%% file: tables/app_decoder_controls.tex
\begin{table}[!htbp]
\centering
\small
\caption{Fixed-input origin control (AP, \%).  Part masks, classes, scores, and axes are identical; only the representative-origin rule changes.}
\label{tab:app-origin-control}
\setlength{\tabcolsep}{5pt}
\begin{tabularx}{\linewidth}{@{}Xrrrr@{}}
\toprule
\tablehead
Origin rule & Part AP$_{50}$ & Axis-gated & Origin-gated & \apmotion \\
\midrule
Centroid (no handle cue) & 47.93 & 43.75 & 14.37 & 13.74 \\
\rowours Handle-guided hinge & 47.93 & 43.75 & 43.11 & \textbf{40.98} \\
\bottomrule
\end{tabularx}
\end{table}

\begin{table}[!htbp]
\centering
\small
\caption{Matched learned heads on two frozen part-model draws (joint line-distance AP, \%).  Brackets are paired 95\% scene-jackknife intervals for learned-head minus training-free-rule AP (pp); each includes zero.}
\label{tab:app-learned-two-bases}
\setlength{\tabcolsep}{3.5pt}
\begin{tabularx}{\linewidth}{@{}Xrr@{}}
\toprule
\tablehead
Decoder & Reference part model & Second part-model draw \\
\midrule
Training-free rule & 40.98 & 39.45 \\
Learned candidate selection & 38.63 [$-5.35,+0.63$] & 38.63 [$-4.54,+2.91$] \\
Learned selection, box axes only & 38.78 [$-5.33,+0.92$] & 39.39 [$-4.44,+4.31$] \\
Rule axis + learned edge & 39.36 [$-4.31,+1.05$] & 37.64 [$-4.66,+1.04$] \\
\bottomrule
\end{tabularx}
\vspace{2pt}
\begin{minipage}{\linewidth}\footnotesize
Each learned row is the best of three head seeds by 42-scene validation motion AP for its backbone, with predicted-handle evidence. Checkpoints within each seed are selected by held-out training-fold loss. This compares frozen-feature decoders, not jointly trained representations.
\end{minipage}
\end{table}

%% file: tables/app_coverage.tex
\begin{table}[!htbp]
\centering
\small
\caption{Coverage and first-failure diagnostics on validation.  Candidate coverage and selection are not end-to-end AP.}
\label{tab:app-coverage}
\setlength{\tabcolsep}{4pt}
\begin{tabularx}{\linewidth}{@{}Xrcc@{}}
\toprule
\tablehead
Population / diagnostic & $n$ & Coverage & Selection given coverage \\
\midrule
Movable parts matched by predicted mask & 390 & 64.4\% & -- \\
Matched rotations: passing hinge candidate & 1,423 & 94.4\% & 88.3\% \\
GT rotations: passing hinge candidate & 236 & 95.8\% & 88.1\% \\
GT handles matched by child union & 388 & 49.7\% & -- \\
\bottomrule
\end{tabularx}

\vspace{5pt}
\begin{tabularx}{\linewidth}{@{}Xrrr@{}}
\toprule
\tablehead
First movable-part failure & Rotation & Translation & Total \\
\midrule
No mask with IoU $>0.5$ & 50 & 89 & 139 \\
Covered; axis gate fails & 12 & 2 & 14 \\
Covered; axis passes; origin fails & 14 & 0 & 14 \\
All motion gates pass & 160 & 63 & 223 \\
\midrule
GT population & 236 & 154 & 390 \\
\bottomrule
\end{tabularx}
\vspace{2pt}
\begin{minipage}{\linewidth}\footnotesize
The handle-union coverage by ground-truth-size quartile is 22/97, 53/99, 63/96, and 55/96, from smallest to largest.  Counts are descriptive and do not represent additive AP gains.
\end{minipage}
\end{table}

%% file: figures/final_coverage_selection.tex
\begin{figure}[H]
\centering
\begin{tikzpicture}[inner sep=0pt,outer sep=0pt]
\node[inner sep=0pt,outer sep=0pt] (image) {\includegraphics[width=\linewidth,height=.45\textheight,keepaspectratio]{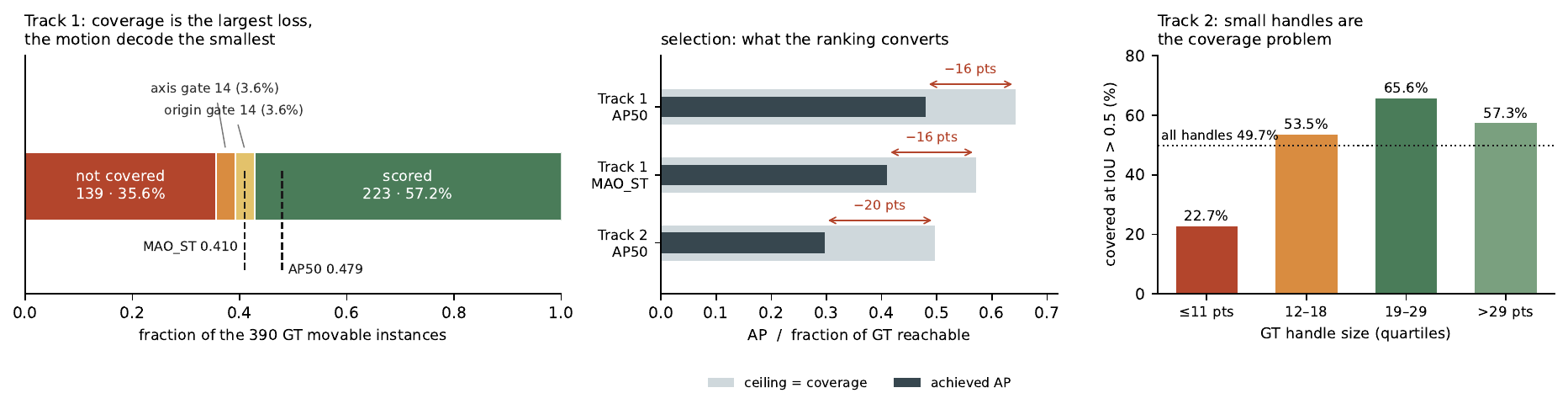}};
\begin{scope}[shift={(image.south west)},x={($(image.south east)-(image.south west)$)},y={($(image.north west)-(image.south west)$)}]
  \fill[white] (0,.875) rectangle (.385,1);
  \fill[white] (.405,.875) rectangle (.705,1);
  \fill[white] (.735,.875) rectangle (1,1);
  \node[font=\sffamily\fontsize{5.7}{6.8}\selectfont,align=center,text=black] at (.19,.942) {Movable-part coverage};
  \node[font=\sffamily\fontsize{5.7}{6.8}\selectfont,align=center,text=black] at (.555,.942) {Coverage and AP};
  \node[font=\sffamily\fontsize{5.7}{6.8}\selectfont,align=center,text=black] at (.867,.942) {Handle coverage};
  \foreach \y/\t in {.748/{Part\\AP50},.575/{Part\\motion AP},.405/{Handle\\AP50}} {
    \fill[white] (.376,\y-.047) rectangle (.421,\y+.047);
    \node[font=\sffamily\fontsize{4.25}{4.8}\selectfont,align=center,text=black] at (.399,\y) {\t};
  }
  \fill[white] (.43,0) rectangle (.77,.105);
  \node[font=\sffamily\fontsize{5.5}{6.5}\selectfont,align=center,text=black!75] at (.60,.045) {coverage fraction and AP\\shown on a common 0--1 scale};
\end{scope}
\end{tikzpicture}
\caption{\textbf{Coverage and selection limits on the public validation split.} Left: the movable-part census assigns each ground-truth instance to coverage, axis, origin, or fully scored outcomes. Centre: each metric's macro AP is juxtaposed with a pooled instance-coverage fraction on a common 0--1 scale. These coverage fractions are diagnostic reachability statistics, not strict AP ceilings or quantities subtractable from AP. Right: handle coverage by ground-truth size quartile.}
\label{fig:appendix-coverage-selection}
\end{figure}

%% file: figures/qualitative_baseline.tex
\begin{figure}[H]
\centering
\begin{tikzpicture}[inner sep=0pt,outer sep=0pt]
\node[inner sep=0pt,outer sep=0pt] (image) {\includegraphics[width=\linewidth,trim=0bp 203bp 0bp 134bp,clip]{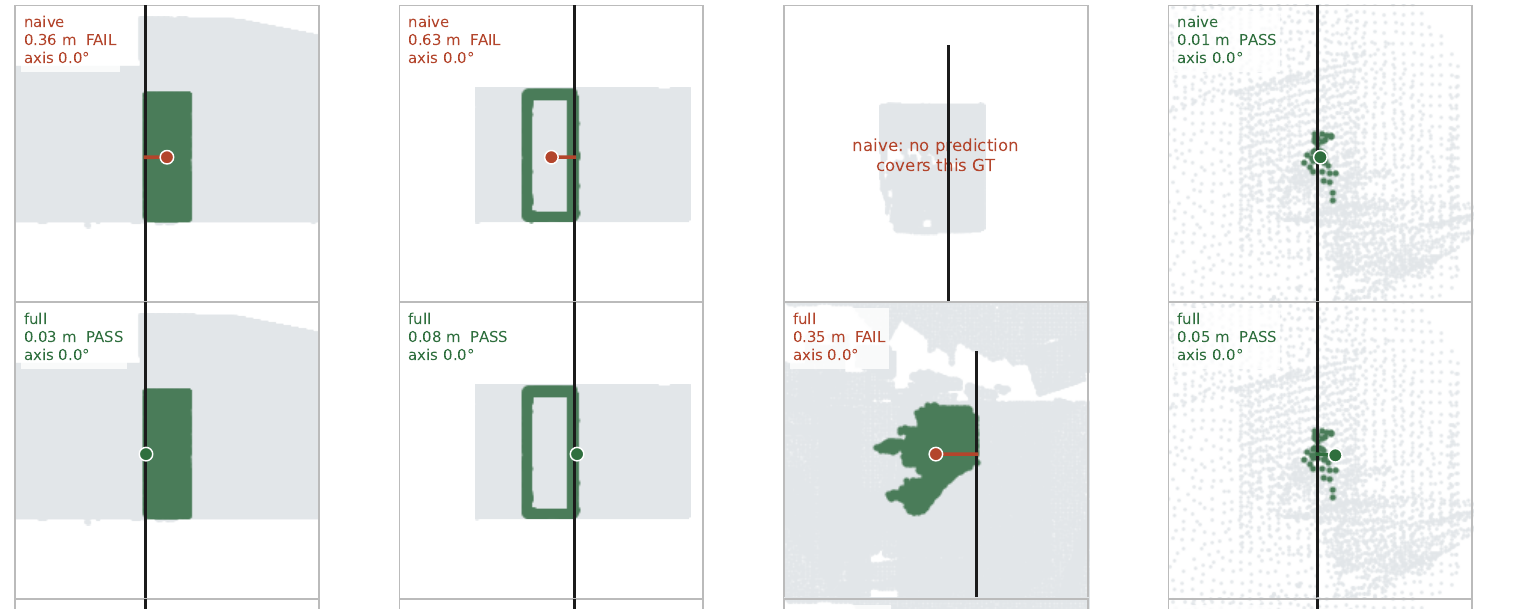}};
\begin{scope}[shift={(image.south west)},x={($(image.south east)-(image.south west)$)},y={($(image.north west)-(image.south west)$)}]
\foreach \x/\t in {0/{Median naive-\\origin error},.25/{90th-percentile\\naive-origin error},.50/{Full-model-only\\coverage},.75/{Naive origin\\is better}} {
  \node[anchor=south,font=\sffamily\fontsize{6.5}{7.5}\selectfont,align=center,text width=.235\linewidth,text=black!75] at (\x+.125,1.025) {\t};
}
\end{scope}
\end{tikzpicture}
\caption{\textbf{Naive versus full geometric decoding.} The top and bottom rows compare the naive and full decoders against the same ground-truth part; black denotes the annotated hinge. The first two columns are naive-error ranks 92 and 165 among 183 rotations covered by both configurations. The third is the median-size instance among three rotations covered only by the full configuration; its origin still fails. The last has the smallest naive-origin error among 38 both-pass cases, and the naive origin is closer. No translation is covered by only one configuration in the selection population. The gallery retains these exceptions rather than estimating their frequency.}
\label{fig:qual-baseline}
\end{figure}

%% file: figures/final_weak_handle_field.tex
\begin{figure}[H]
\centering
\includegraphics[width=\linewidth,height=.70\textheight,keepaspectratio,trim=0bp 118bp 0bp 74bp,clip]{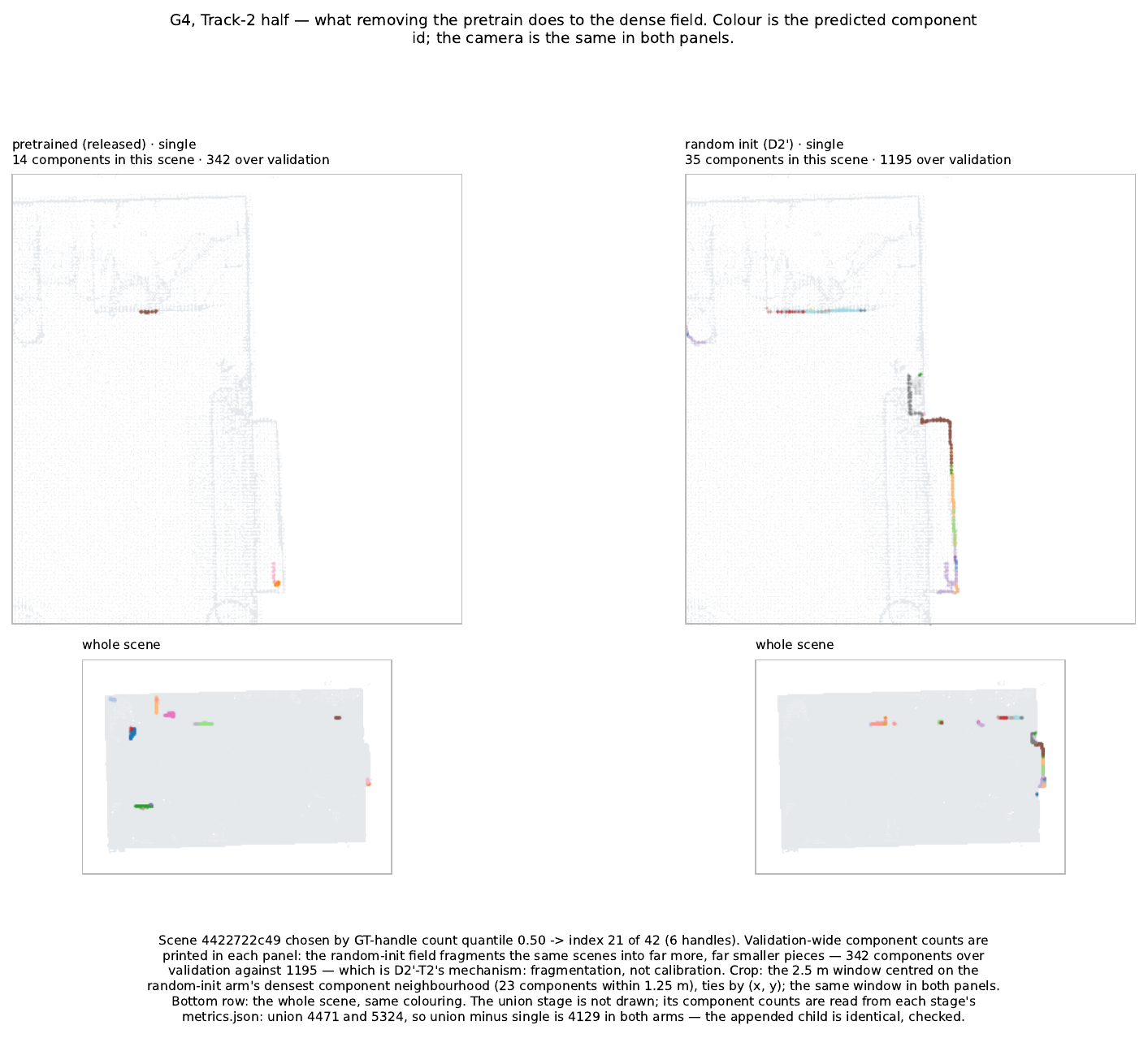}
\caption{\textbf{Dense-handle fields under source variation.} The same validation crop and camera compare the reference and random-initialization configurations from \Cref{tab:app-source-variation}. Their training schedules also differ, so the comparison does not isolate pretraining. The scene has the median ground-truth-handle count (21st of 42 scenes). The random-initialized field emits more fragmented components (35 versus 14 in this scene; 1,195 versus 342 over validation). These are dense predictions before child augmentation.}
\label{fig:appendix-weak-handle-field}
\end{figure}

%% file: figures/appendix_galleries.tex
\begin{figure}[!htbp]
\centering
\includegraphics[width=\linewidth,trim=7bp 36bp 579bp 74bp,clip]{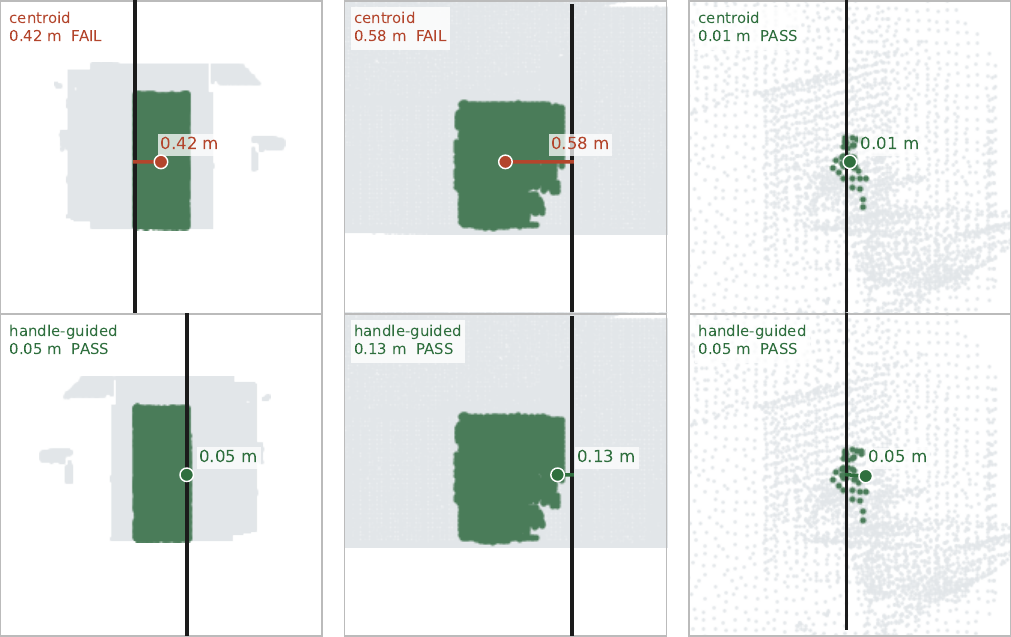}
\par\medskip
\includegraphics[width=\linewidth,trim=503bp 36bp 82bp 74bp,clip]{appendix/qualitative_hinge_source.pdf}
\caption{\textbf{Hinge successes and failures.} Within each pair, the top row uses a centroid origin and the bottom row uses handle guidance for the same predicted part. Upper group, left to right: centroid-error ranks 61 and 109 among 121 rotations that pass with guidance but fail with centroid origins, and the smallest centroid error among 37 both-pass cases. Lower group: median axis error among 12 non-vertical hinges, median origin error among 19 missing-handle fallbacks, and the largest guided-origin error among 186 covered rotations. Black lines denote annotated hinges. No case in this 186-instance population passes with the centroid but fails with guidance; the last column is a residual failure, not a reversal example.}
\label{fig:app-hinges}
\end{figure}

\begin{figure}[!htbp]
\centering
\newcommand{\componentview}[4]{\includegraphics[width=.25\linewidth,viewport=#1 #2 #3 #4,clip]{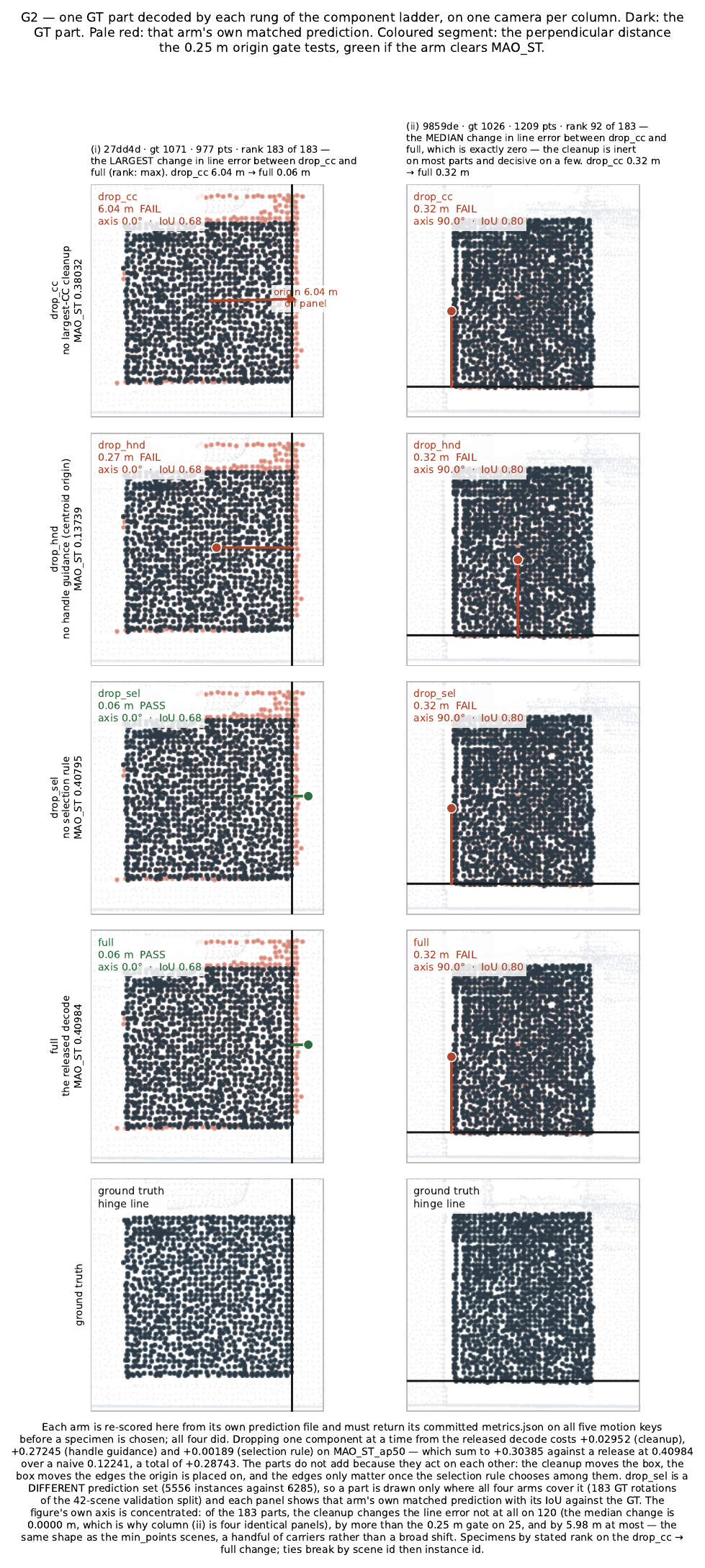}}
\setlength{\tabcolsep}{3pt}
\begin{tabular}{@{}ccc@{}}
\small No support cleanup & \small Centroid origin & \small Full decoder \\[4pt]
\multicolumn{3}{@{}l}{\small Largest cleanup-induced change (rank 183/183)} \\[3pt]
\componentview{60}{766}{216.3}{922.5} &
\componentview{60}{600.5}{216.3}{756.8} &
\componentview{60}{269.1}{216.3}{425.4} \\
\small 6.04\,m & \small 0.27\,m & \small 0.06\,m \\[7pt]
\multicolumn{3}{@{}l}{\small Median cleanup-induced change (rank 92/183)} \\[3pt]
\componentview{270.5}{766}{426.8}{922.5} &
\componentview{270.5}{600.5}{426.8}{756.8} &
\componentview{270.5}{269.1}{426.8}{425.4} \\
\small 0.32\,m & \small 0.32\,m & \small 0.32\,m
\end{tabular}
\caption{\textbf{Where geometric components matter.} Rows fix the part and camera. Dark points are ground truth, pale red the prediction, and black the annotated hinge; values are perpendicular origin errors. The upper case has the largest cleanup-induced change among 183 jointly covered rotations (its uncleaned origin is outside the view). The median-change case below is unchanged and has a horizontal hinge, violating the upright-axis prior.}
\label{fig:app-components}
\end{figure}

\begin{figure}[!htbp]
\centering
\newcommand{\failureview}[4]{\includegraphics[width=.22\linewidth,viewport=#1 #2 #3 #4,clip]{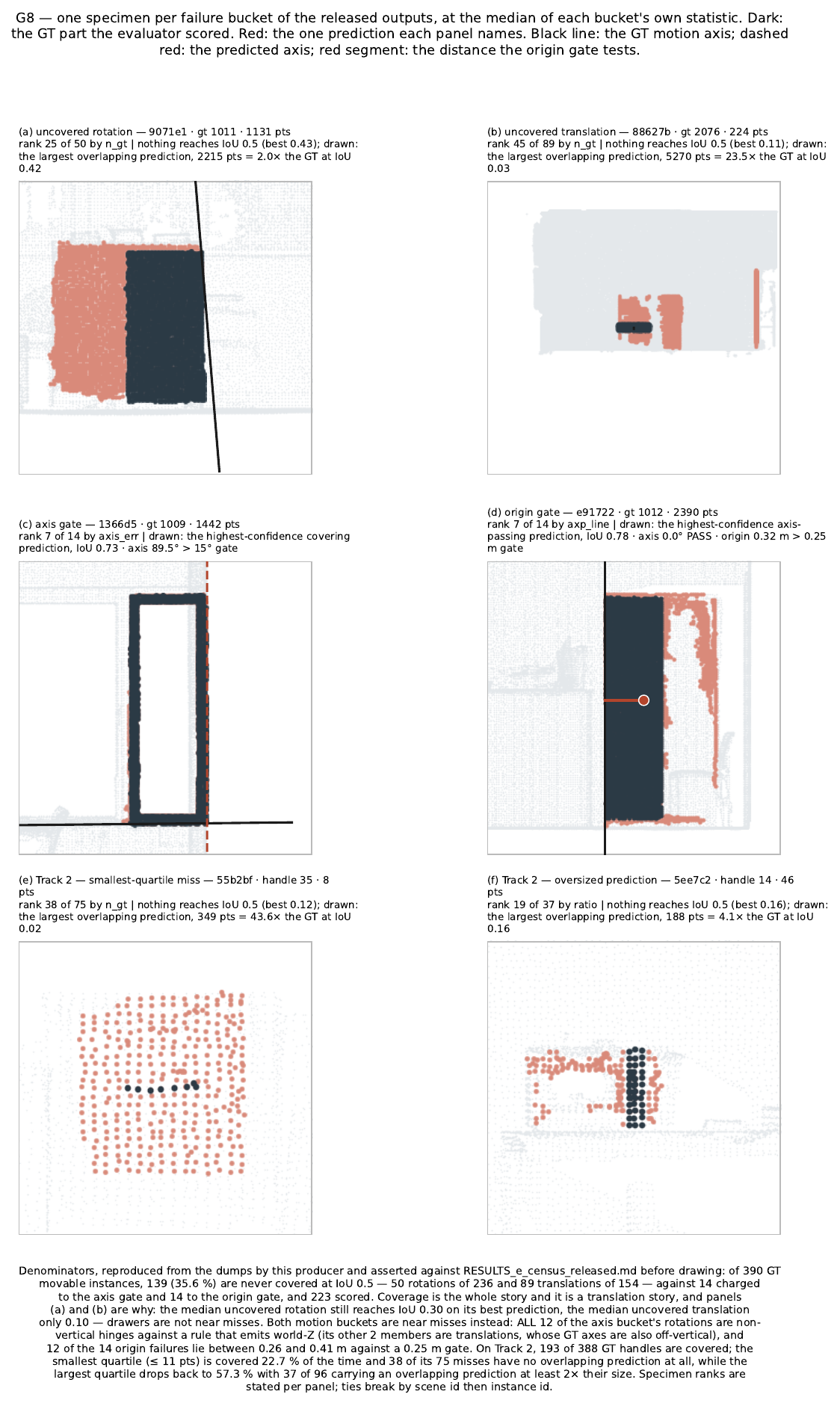}}
\setlength{\tabcolsep}{3pt}
\begin{tabular}{@{}ccc@{}}
\small Uncovered rotation & \small Uncovered translation & \small Axis failure \\[3pt]
\failureview{11.6}{597.5}{201}{786.9} & \failureview{312.5}{597.5}{501.9}{786.9} & \failureview{11.6}{353.4}{201}{542.8} \\[7pt]
\small Origin failure & \small Small-handle miss & \small Over-extended handle \\[3pt]
\failureview{312.5}{353.4}{501.9}{542.8} & \failureview{11.6}{109.2}{201}{298.7} & \failureview{312.5}{109.2}{501.9}{298.7}
\end{tabular}
\caption{\textbf{Distinct sources of error.} Dark points are ground truth, pale red the prediction, and black the annotated hinge. Median-rank cases show uncovered rotations (25/50), uncovered translations (45/89), axis error (7/14), origin error (7/14), smallest-quartile handle misses (38/75), and over-extension (19/37). Frequencies come from the census, not this gallery.}
\label{fig:app-failures}
\end{figure}

\begin{figure}[!htbp]
\centering
\begin{minipage}[t]{.49\linewidth}
\centering\small Median reference-origin error\par\smallskip
\includegraphics[width=.80\linewidth,trim=18bp 56bp 361bp 114bp,clip]{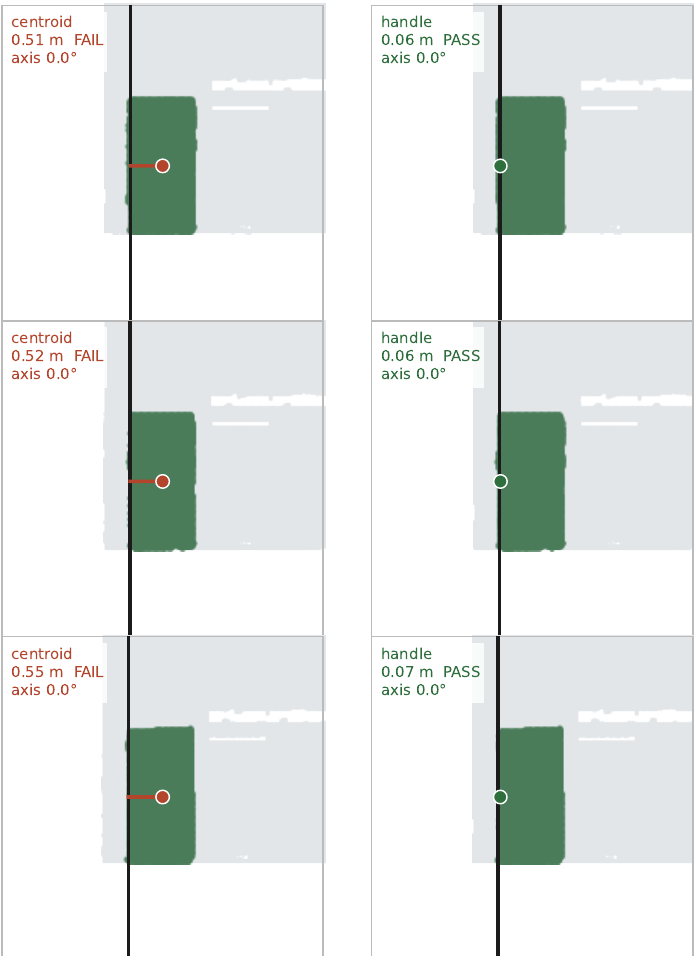}
\end{minipage}\hfill
\begin{minipage}[t]{.49\linewidth}
\centering\small 90th-percentile reference-origin error\par\smallskip
\includegraphics[width=.80\linewidth,trim=372bp 56bp 5bp 114bp,clip]{appendix/qualitative_weaker_parts_source.pdf}
\end{minipage}
\caption{\textbf{Handle guidance with different part predictors.} Rows use reference, joint-model, and random-initialization parts. Each pair changes only the centroid origin (left) to dense-handle guidance (right); black marks the annotated hinge. Specimens have reference-origin-error ranks 61 and 110 among 122 rotations covered by all three sources. Guidance helps across sources. Random initialization also changes training duration and does not isolate pretraining.}
\label{fig:app-weaker-parts}
\end{figure}

\begin{figure}[!htbp]
\centering
\includegraphics[width=.76\linewidth,trim=100bp 51bp 676bp 76bp,clip]{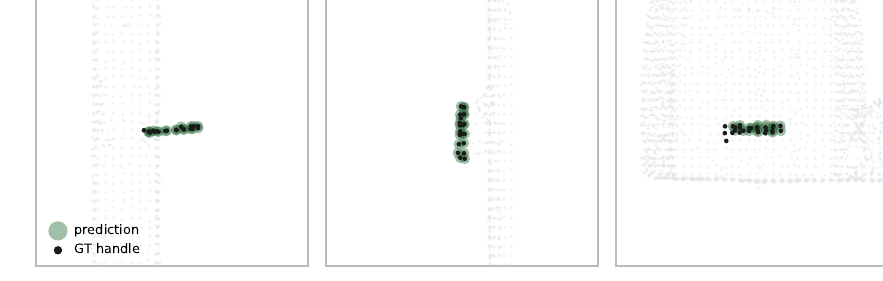}
\par\smallskip\small Dense success (median size)\hfill Dense success (above median)\hfill Child-only recovery
\par\medskip
\includegraphics[width=.76\linewidth,trim=533bp 51bp 100bp 76bp,clip]{appendix/qualitative_handles_source.pdf}
\par\smallskip\small Smallest recovery\hfill Over-extended dense\hfill Class correction\hfill Unmatched child
\caption{\textbf{Handle proposal and class behavior.} Black points are annotated handles; colors show predictions. The top row includes two dense successes and the median-size child-only recovery. Below: the smallest recovery, largest dense over-extension ratio, matched class correction, and an unmatched corrected child. Correction cases have median annotated size in their respective strata. Selections are deterministic, not random.}
\label{fig:app-handles}
\end{figure}

%% file: sections/07_external_comparison.tex
\section{External System Comparison}
\label{app:external}

We run REACT3D as an external baseline for articulated-part output on the same 42 validation scenes. Unlike the published-number comparison, REACT3D is evaluated by us with the same evaluator as \method. Its pipeline consumes richer inputs---RGB-D keyframes, camera poses, and a scan mesh---rather than our RGB point cloud. Both outputs are expressed as part instances with a motion type, axis, and origin. Input and training conditions are not controlled comparison axes: REACT3D is not trained on Articulate3D, whereas our perception models use its training split. Conversely, 20 of the 42 validation scenes occur in REACT3D's authors' development lists and therefore cannot be described as unseen during its development.

We use REACT3D's released processing stages with compatibility changes required by the available hardware and data representation. A PyTorch API-compatibility patch to its MSDeformAttn dispatch preserves the kernel and numeric type. Data preparation follows the released image format and frame schedule. For evaluation, reconstructed part surfaces are transferred to the evaluation point cloud by a reverse nearest-neighbour rule: a cloud point belongs to a part when its nearest part vertex lies within 0.02\,m. This expresses predictions on the point-labelled evaluation domain. We normalize every predicted axis before scoring because raw REACT3D axis vectors are not unit length; without normalization, the evaluator's origin projection is invalid even though its axis-angle test is unchanged. These compatibility and conversion steps do not imply byte-identical execution; the released reference output was checked for compatible part, class, axis, and origin behavior.

All 42 scenes were scored; none was omitted. Three scenes completed the pipeline but produced no accepted part; their stage logs certify empty prediction sets, which enter the evaluator accordingly. REACT3D produces 199 parts for 390 annotated parts. The accuracy comparison in \Cref{tab:react3d-accuracy} and diagnostic in \Cref{tab:app-react3d} show that its low translation result is principally a coverage issue: only 7 of 154 translation parts have a same-class mask match, while 98 are untouched and 48 are touched but below the IoU gate. All seven matched translations pass the axis gate. For rotations, 18 of 61 mask-matched parts fail only the origin gate, separating residual motion error from the larger detection deficit. The paired scene comparison is descriptive of this fixed run: \method has higher motion AP on 40 of the 41 scenes containing a ground-truth part and ties on one; one scene contains no ground-truth part.

\input{tables/app_react3d}

\paragraph{Timing and memory accounting}
The measurements in \Cref{tab:react3d-cost} use one RTX 5070 Ti and 30\,GB of host RAM, one scene at a time, from each system's declared input through its articulated-part predictions. Time sums each stage's last successful attempt, including model loading; failed and superseded attempts are excluded. REACT3D input starts at extracted keyframes, poses, and mesh, so video decoding and input-frame checks are excluded as dataset preparation. Our superpoint extraction is included and uses no additional worker parallelism. This measures a complete inference pass, not the elapsed duration of the experimental campaign.

REACT3D's 22.94\,GiB as-shipped host-memory peak comes from its original point-assignment stage, used on 22 scenes. The other 20 use an output-identical memory refit with a separately measured peak of 16.04\,GiB. Its GPU logger began partway through the run: 8,780\,MiB is the maximum over nine logged scenes, whose peaks span 7,616--8,780\,MiB; the other 33 scenes were not GPU-logged. We do not extrapolate these measurements. Our 10,262\,MiB \texttt{nvidia-smi} peak covers all 42 scenes at 0.2\,s polling; its separately instrumented allocated-memory peak is 7.17\,GiB. The unequal logging coverage precludes a matched GPU-efficiency conclusion.

The \method timing path includes superpoints, dense handles, standalone parts, and motion decoding; it excludes the joint predictor needed only for additional handle outputs. Checkpoint-derived parameter counts cover these two models (200,610,183 parameters), versus five REACT3D models (5,397,612,242). The cost comparison in \Cref{tab:react3d-cost} consequently concerns articulated-part output, not the complete three-predictor interaction system. The timed \method path reproduces reference validation masks, ordering, classes, axes, and origins on all 42 scenes; scores differ by at most $9.39\times10^{-6}$, with no effect on any evaluated quantity. The source of this numerical residue is unresolved.

\paragraph{Figure selection and complete-scene reporting}
The main comparison uses the first two scenes under a deterministic ranking: the advantage in matched true positives among displayed predictions, restricted to scenes with at least five annotated parts, with ties broken by motion-AP margin. Our display threshold is $s\geq0.50$; all REACT3D parts are displayed. \Cref{fig:react3d-showcase-appendix} adds the next two ranked scenes. This is favorable-case selection, not a representative sample. \Cref{fig:react3d-scenes} reports every validation scene, including the three certified empty prediction sets and the scene without ground-truth parts. \Cref{fig:react3d-efficiency} summarizes the measured accuracy--time trade-off under the timing and memory scopes above.

\input{figures/react3d_appendix}

%% file: tables/app_react3d.tex
\begin{table}[!htbp]
\centering
\small
\caption{REACT3D coverage and motion diagnostics on all 42 validation scenes. The four mask-outcome rows partition the ground-truth parts in each motion class. Counts describe coverage, not AP contributions.}
\label{tab:app-react3d}
\setlength{\tabcolsep}{5pt}
\begin{tabularx}{\linewidth}{@{}Xrr@{}}
\toprule
\tablehead Outcome & Rotation & Translation \\
\midrule
Ground-truth parts & 236 & 154 \\
Predicted parts & 155 & 44 \\
\midrule
Same-class mask match (IoU $>0.5$) & 61 & 7 \\
No overlapping prediction & 158 & 98 \\
Overlap below the mask gate & 16 & 48 \\
Above-gate overlap, wrong class & 1 & 1 \\
\midrule
Mask-matched parts passing motion gates & 39 & 7 \\
\bottomrule
\end{tabularx}
\vspace{2pt}
\begin{minipage}{\linewidth}\footnotesize
The below-gate row combines weak overlap and near-miss categories. Among 61 mask-matched rotations, 18 fail only the origin gate and four fail both motion gates. All seven matched translations pass the axis gate. The prediction counts sum to 199; the annotation counts sum to 390.
\end{minipage}
\end{table}

%% file: figures/react3d_appendix.tex
\begin{figure}[!htbp]
\centering
\setlength{\tabcolsep}{1.5pt}
\begin{tabular}{@{}cccc@{}}
\small RGB & \small GT + axes & \small \method & \small REACT3D \\[2pt]
\includegraphics[width=.237\linewidth]{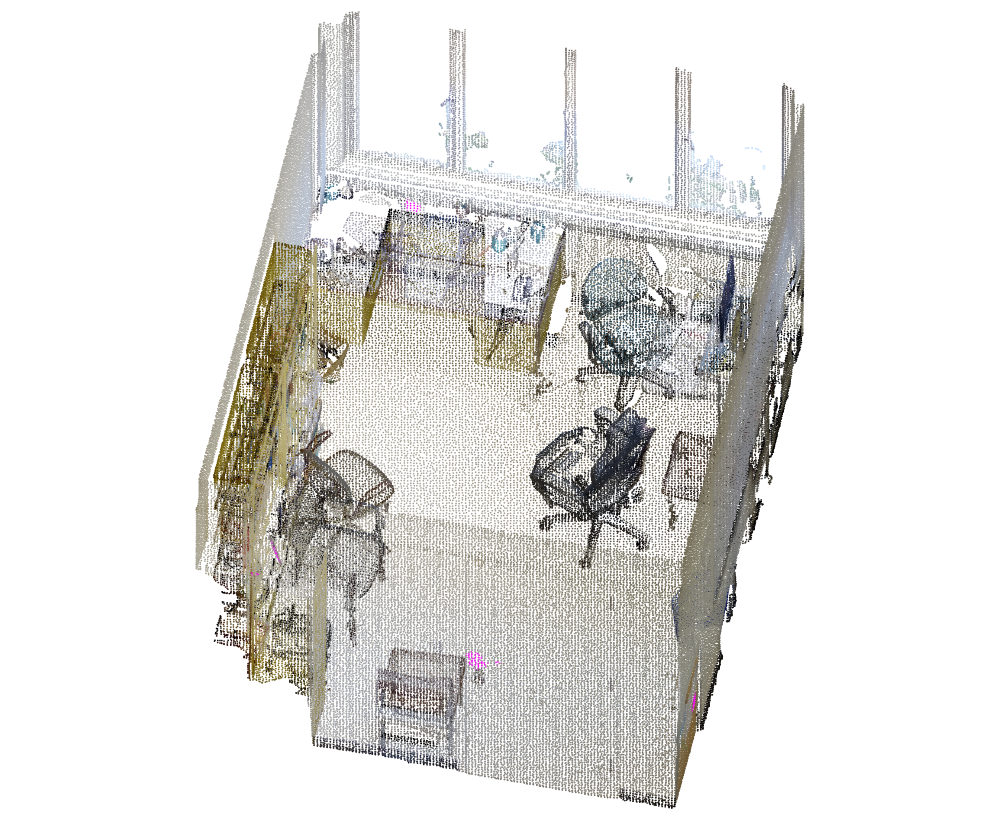} &
\includegraphics[width=.237\linewidth]{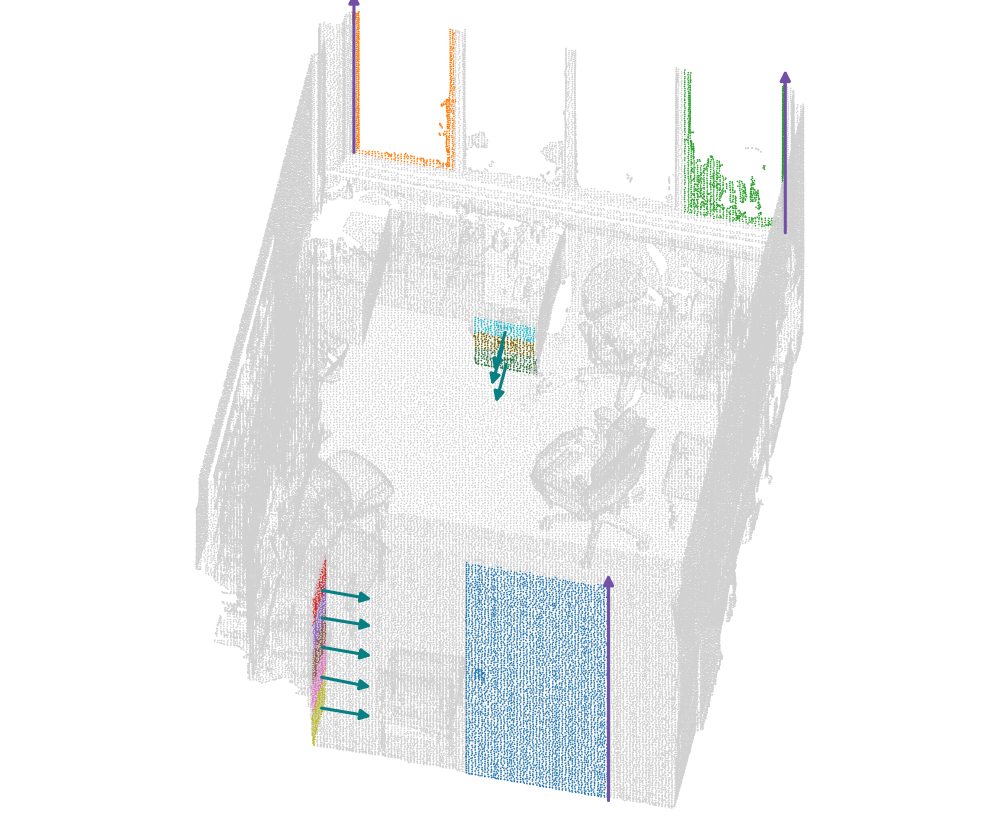} &
\includegraphics[width=.237\linewidth]{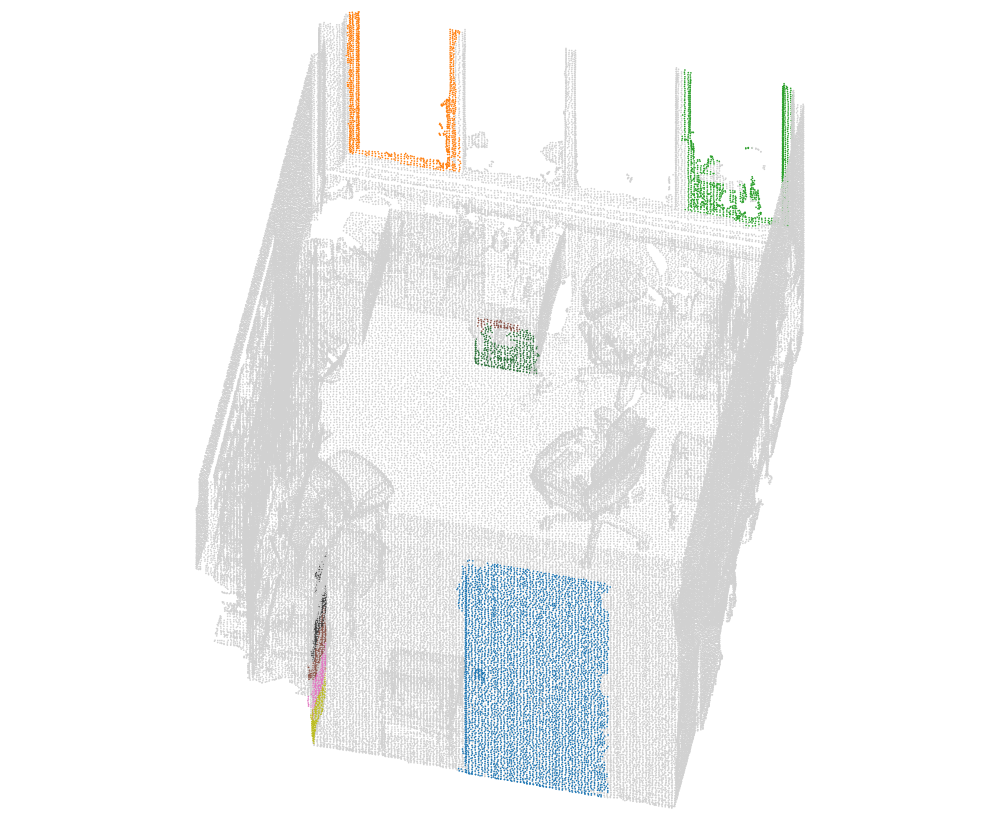} &
\includegraphics[width=.237\linewidth]{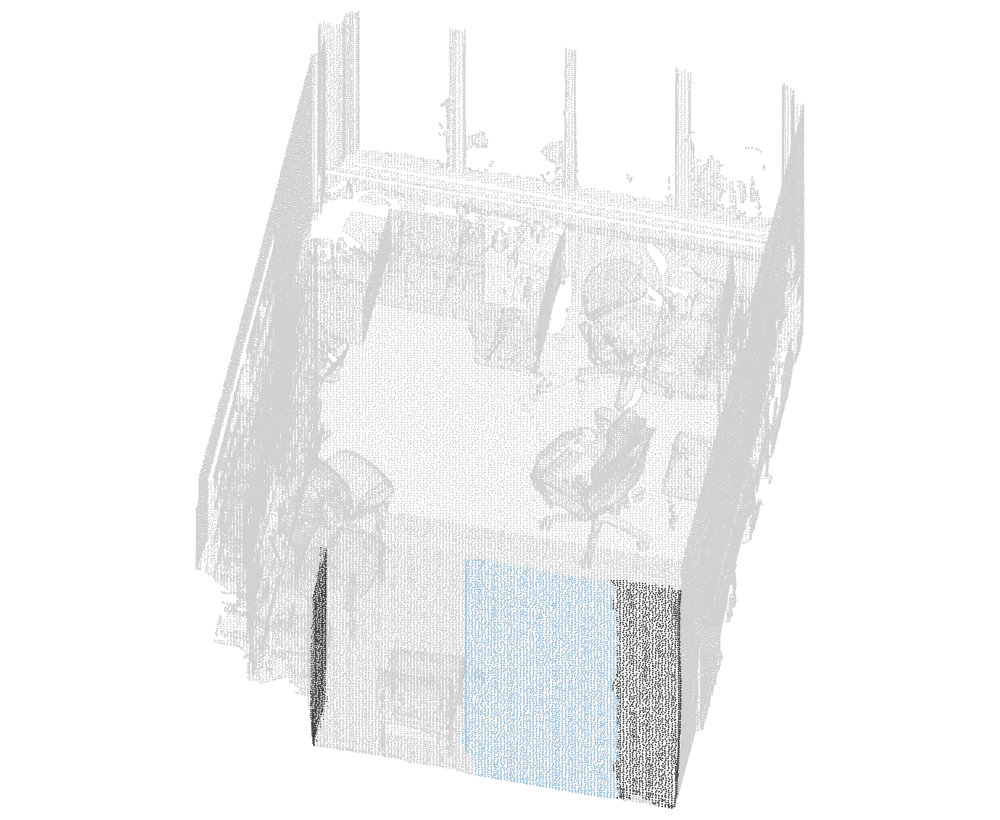} \\
\multicolumn{4}{c}{\scriptsize 8a20d62ac0: 11 GT; ours 7 matched of 8 shown; REACT3D 0 matched of 3} \\[1pt]
\includegraphics[width=.237\linewidth]{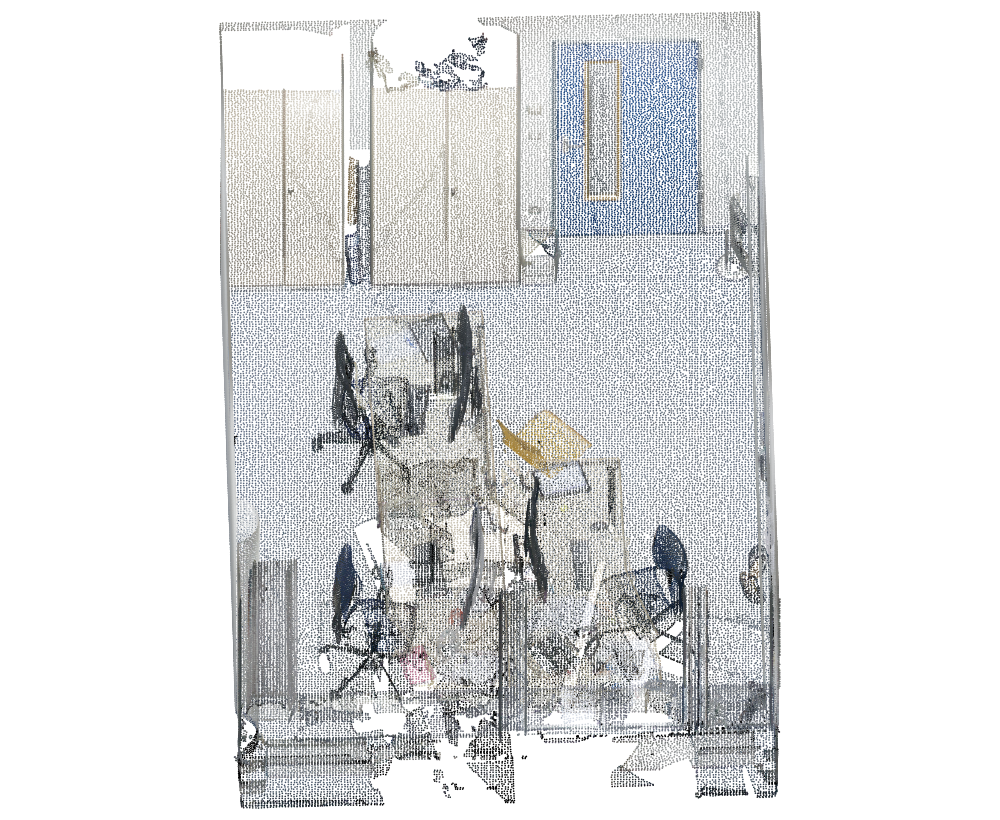} &
\includegraphics[width=.237\linewidth]{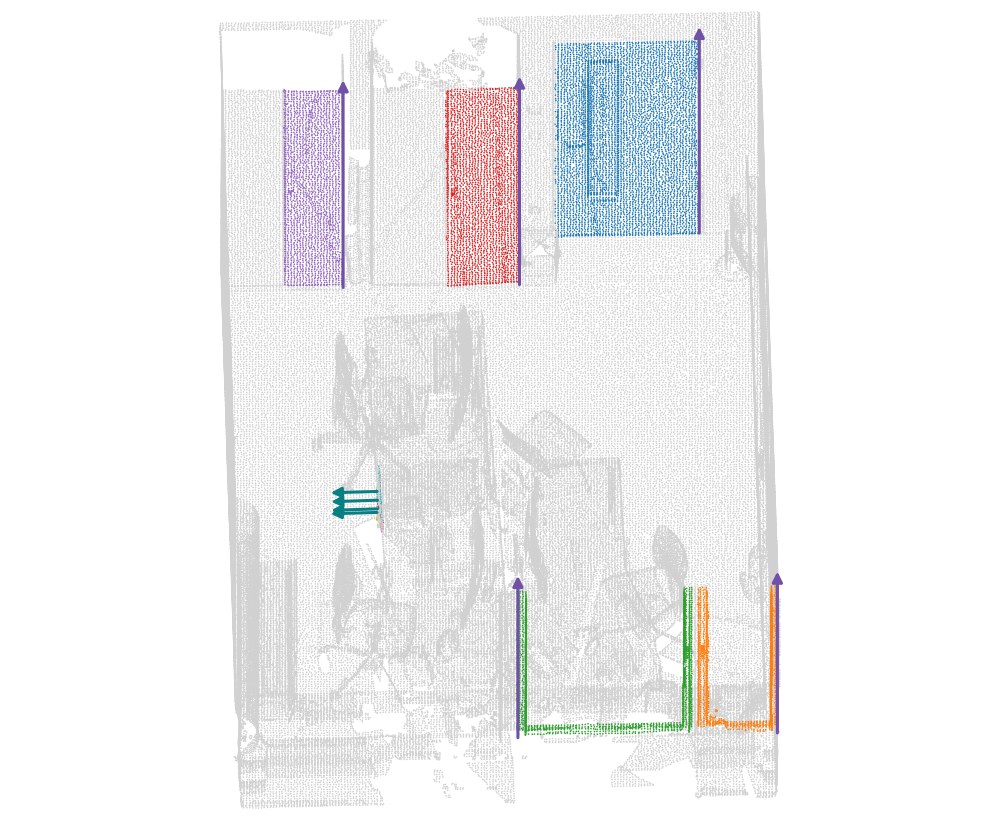} &
\includegraphics[width=.237\linewidth]{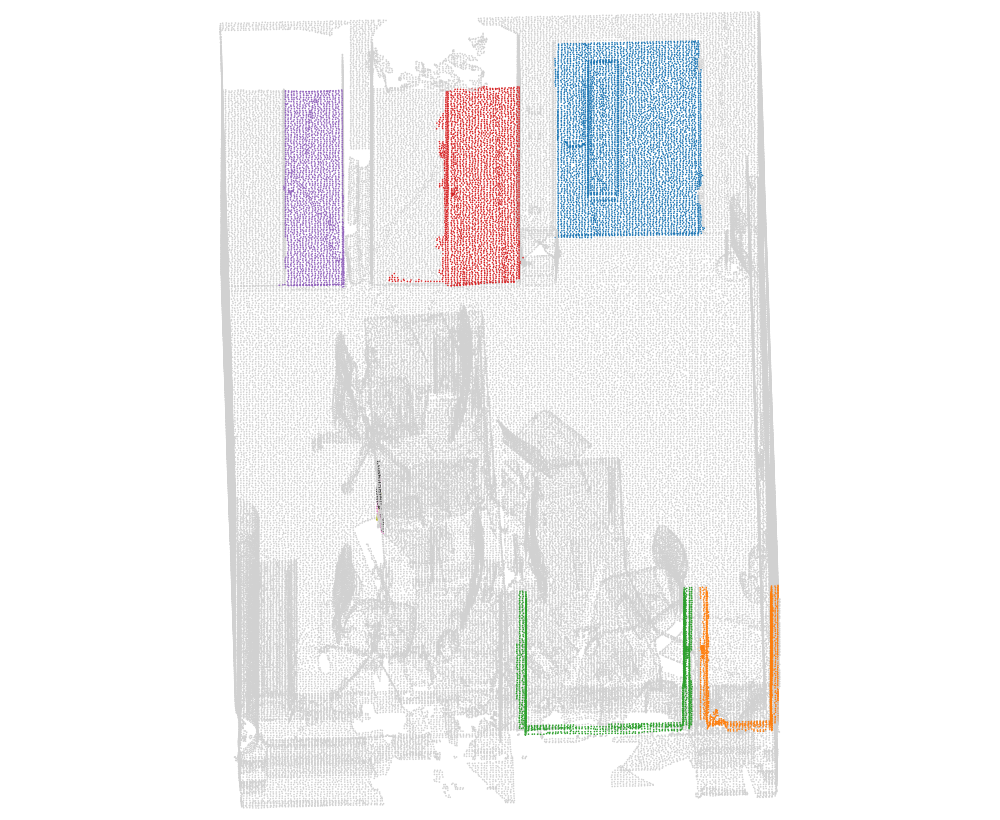} &
\includegraphics[width=.237\linewidth]{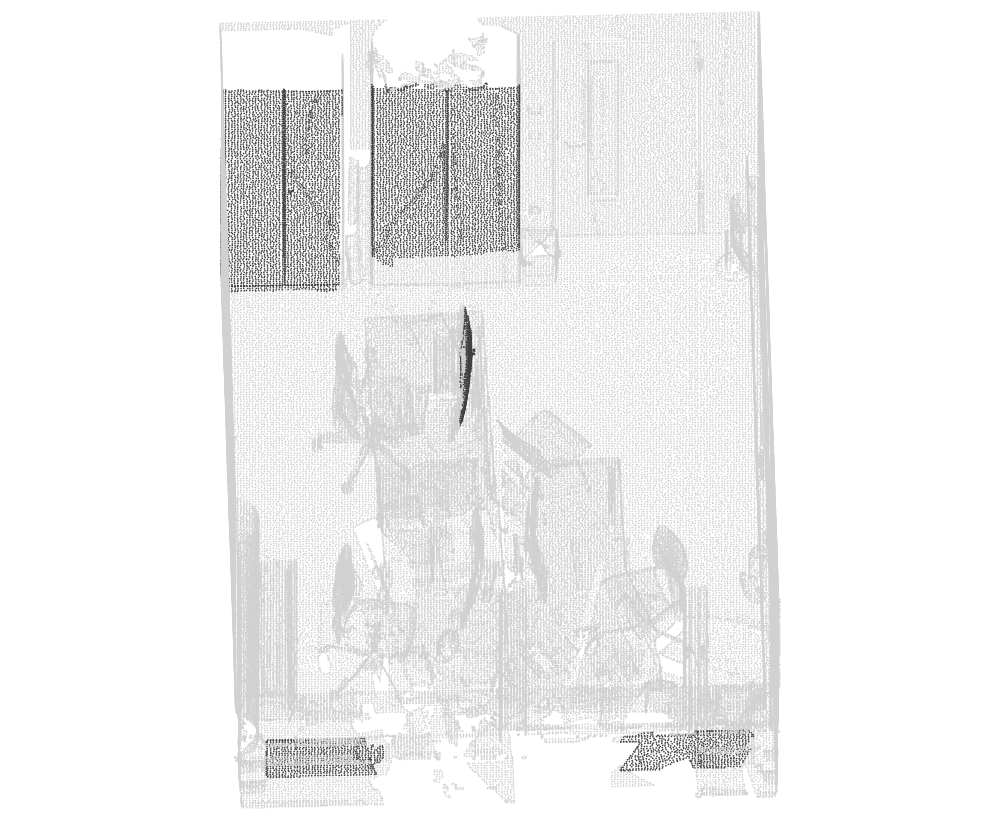} \\
\multicolumn{4}{c}{\scriptsize 4a1a3a7dc5: 9 GT; ours 7 matched of 9 shown; REACT3D 0 matched of 7}
\end{tabular}
\caption{\textbf{Additional qualitative comparison with REACT3D.} The next two scenes under the ranking used in the main report: at least five annotated parts, ordered by displayed true-positive advantage and then motion-AP margin. Our threshold is $s\geq0.50$; all REACT3D predictions are shown. Saturated colors indicate correct masks and motion, pale colors indicate mask matches that fail a motion gate, and gray indicates unmatched detections. Purple and teal axes denote rotation and translation. Input and training conditions are given in \Cref{app:external}.}
\label{fig:react3d-showcase-appendix}
\end{figure}

\begin{figure}[!htbp]
\centering
\includegraphics[width=\linewidth]{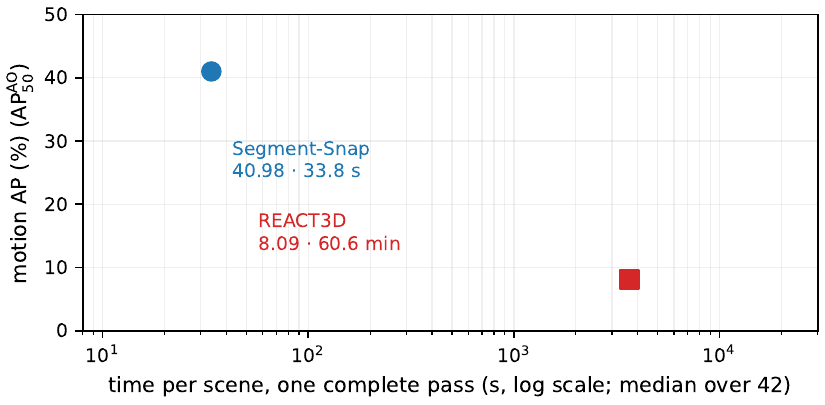}
\caption{\textbf{Motion accuracy and inference time.} Full-validation motion AP versus median complete-pass time per scene (log scale), including model loading on the same RTX 5070 Ti. \method's two-model part-motion path and REACT3D produce common outputs; input, training, and timing scopes are specified in \Cref{app:external}.}
\label{fig:react3d-efficiency}
\end{figure}

\begin{figure}[!htbp]
\centering
\includegraphics[width=\linewidth]{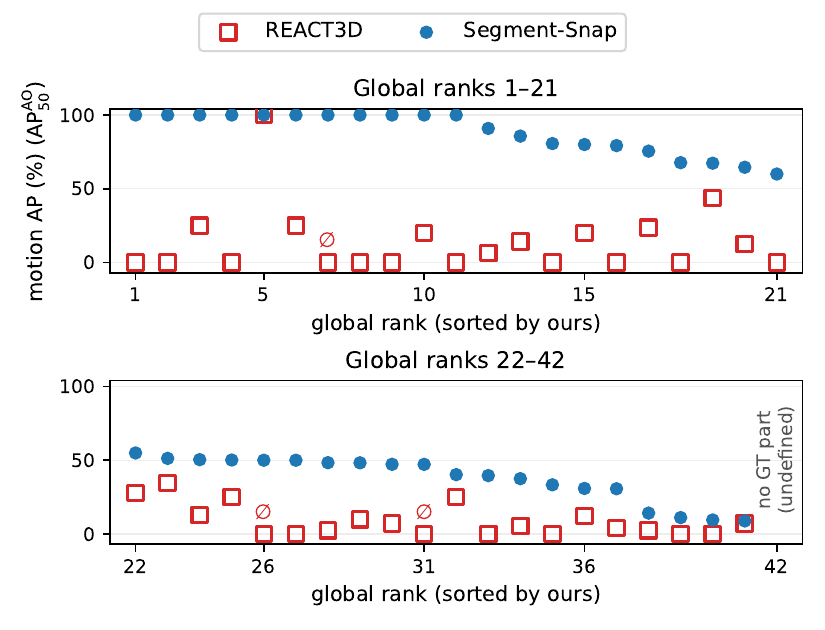}
\caption{\textbf{Per-scene motion AP on all 42 validation scenes.} The 41 scenes with defined AP are sorted by \method's score; $\varnothing$ marks three verified empty REACT3D outputs. The scene without an annotated part (a980334473) is appended at rank 42: its AP is undefined, not zero. \method has higher AP on 40 scored scenes and ties on one; evaluation conditions are specified in \Cref{app:external}.}
\label{fig:react3d-scenes}
\end{figure}